\documentclass[12pt]{article}
\usepackage{iftex}
\ifPDFTeX
\usepackage[T1]{fontenc}
\usepackage{lmodern}
\else
\usepackage{fontspec}
\fi
\usepackage[protrusion=true]{microtype}
\usepackage{geometry}
\usepackage{graphicx}
\usepackage{amsmath}
\usepackage{booktabs}
\usepackage{siunitx}
\usepackage[dvipsnames]{xcolor}
\usepackage[hidelinks]{hyperref}
\usepackage{url}
\usepackage{setspace}
\usepackage[font=small,labelfont=bf,skip=6pt]{caption}
\usepackage{titlesec}
\titlespacing*{\section}{0pt}{1.2em}{0.6em}
\titlespacing*{\subsection}{0pt}{1em}{0.5em}
\titlespacing*{\subsubsection}{0pt}{0.9em}{0.4em}
\titleformat{\paragraph}[block]{\normalfont\normalsize\bfseries}{}{0pt}{}
\titlespacing*{\paragraph}{0pt}{0.8em}{0.3em}
\usepackage{enumitem}

\definecolor{capblue}{HTML}{2C6FB5}
\definecolor{capred}{HTML}{C0392B}
\definecolor{capgray}{HTML}{7F8C8D}
\definecolor{capgreen}{HTML}{2E8B57}

\ifPDFTeX\else
\usepackage{unicode-math}
\fi

\setlist{leftmargin=2em, itemsep=0.25em, topsep=0.5em, partopsep=0pt}
\newcommand{\tok}[1]{\texttt{#1}}

\begin{document}
\onehalfspacing

\title{EventGPT: Next-Event Pretraining\\on Banking Customer Histories}
\author{Denis Burakov}
\date{November 2026}
\maketitle

\begin{abstract}
\noindent
A model pretrained to predict a customer's next event can estimate the probability of future events without training on their labels and provide a reusable representation for supervised downstream banking tasks. We pretrain EventGPT, a hierarchical causal transformer, by next-token maximum likelihood estimation on the public Multimodal Banking Dataset (MBD-mini: 100,224 clients, 38.8 million transactions and 66.3 million geolocation pings). Keeping one state per event, EventGPT trains in about two hours on one L4 GPU at windows up to 8,192 tokens. From its next-event distribution, with no labels, EventGPT scores five targets of the form ``the client makes a transaction of a given kind for the first time next month'' at ROC-AUC 0.79 to 0.92. Gradient boosting on 114 engineered features needs 10,000 to 300,000 labeled client-months to match it on four targets and stays below it on the fifth. For product prediction, frozen EventGPT embeddings score below the engineered features (mean ROC-AUC 0.787 against 0.808 over five test folds), and a rank-averaged ensemble of EventGPT, profile fields and token counts with the engineered features adds 0.014 to 0.017, of which 0.007 to 0.016 is EventGPT's own. The code and pretrained weights are released.
\end{abstract}

\section{Introduction}\label{sec:intro}

A model pretrained to predict a customer's next event can be useful in two ways. It can answer questions about future events directly from its own predictions and give one representation per customer that supervised downstream banking tasks can reuse. EventGPT\footnote{It is unrelated to EventGPT for event-camera streams \cite{eventgptcam} and to Event Stream GPT for medical records \cite{esgpt}.} is trained by predicting each next event token of a customer's event stream. This paper describes the architecture, training recipe, and measures both uses on a public banking benchmark.

The usual starting point of modeling bank outcomes is a feature table including counts, sums and recencies computed per customer and window that is fed into a machine learning algorithm. Building the table requires business knowledge (which field is the amount, which event types to include, which windows), and that work is repeated for every new event source. Each new question also needs labels and a model of its own. When events of many kinds with different fields are combined, as in a lender's applications, repayments and account charges, there is often no obvious table to build. A pretrained event model needs neither the table nor (for questions about events) the labels.

Several financial institutions describe foundation models of this kind. Nubank's nuFormer is a flat
causal decoder over customer transaction tokens \cite{nuformer}, Revolut's PRAGMA is a masked model with separate encoders for profile state, single events and history \cite{pragma}, and Visa's TransactionGPT encodes the aspects of each transaction before modeling across transactions \cite{transactiongpt}. Starling Bank has announced a transactional foundation model trained on 15 billion of its transactions, to be deployed in 2027 \cite{starling}. None of these releases checkpoints or evaluates on publicly available data. Even if a checkpoint were released, however, it would be of little use to another institution: its vocabulary is the company's own event nomenclature, so each institution pretrains its own backbone (Section~\ref{sec:per-institution}). EventGPT combines elements of existing approaches: a per-event encoder, as in PRAGMA and TransactionGPT, with the causal pretraining objective of nuFormer. It is designed for pretraining on consumer-lender product histories. Here, we train and evaluate it exclusively on publicly available data to make the results reproducible.

\paragraph{Contributions}
\begin{enumerate}
\item \textbf{An open reference model}: to our knowledge, the first open-weights generative model for banking event sequences, trained and evaluated on a public benchmark with reproducible baselines. The weights read MBD's schema, so another institution reruns the code on its own events (Section~\ref{sec:per-institution}). The tokenization of MBD-mini, the model, the training and evaluation code and the checkpoints are released (Section~\ref{app:code}).
\item \textbf{A hierarchical architecture for event streams}: an event encoder pools the tokens of each event into one state, a history decoder attends causally over the event states, and a token decoder generates the next event. The last hidden state serves as the customer embedding and incorporates information about the time since the client's last activity. Over a window of $n$ tokens, the token layers attend within events at linear cost $O(n)$, while the history layers attend over approximately $n/8$ events at cost $O((n/8)^2)$ against $O(n^2)$ for attention over tokens (Sections~\ref{sec:model} and~\ref{sec:discussion}).
\item \textbf{Zero-shot scoring of event targets} from the token decoder's distribution over the next event, which scores five first-occurrence targets at ROC-AUC 0.79 to 0.92 with no labels. CatBoost on the engineered features reaches the same ROC-AUC after 10,000 to 300,000 randomly sampled client-months on four targets and stays below it on the fifth at 300,000. A Markov chain over event kinds scaled by the client's recent activity, which also needs no labels, scores 0.79 to 0.88. EventGPT scores above it on two of the five targets and on 48 of the 74 kinds with at least 50 test positives, and below it on 7 (Section~\ref{sec:zero-shot}).
\item \textbf{An evaluation of frozen and fine-tuned embeddings on MBD} against engineered features and a schema-free token-count table, with paired standard errors clustered by client. The embedding is below both tables and exceeds token counts only inside its own window. We add the profile fields and full-history token counts to the embedding. The gap then falls to 0.004, averaged over five test folds, and is not statistically significant on any fold. Late fusion of the profile fields adds 0.013 to 0.019 to the ROC-AUC of the product targets and brings \tok{target\_4} level with the engineered features (Sections~\ref{sec:protocol}--\ref{sec:results}).
\item \textbf{The embedding as a second ranker beside a feature table}: averaging its ranks with those of the engineered features adds 0.014 to 0.017 to their ROC-AUC on each of five test folds (intervals excluding zero, weights fixed in advance). The same ensemble built without EventGPT, from the profile fields and token counts alone, adds 0.001 to 0.008, so 0.007 to 0.016 of the gain is EventGPT's own ($p < 0.01$ on each of ten fold and seed pairs). As features inside the table the embedding leaves their ROC-AUC unchanged or lower (Section~\ref{sec:gap}).
\item \textbf{Embeddings as point-in-time features}: an embedding for a date uses only events before
that date, so it can be stored and reused across tasks without leaking the future (Section~\ref{sec:pit}).
\end{enumerate}

\paragraph{Terms}
The \emph{as-of date} is the first day of a month; a customer's history at it is every event before it, and every score and embedding is computed at one. \emph{Product targets}, also called the official targets, are MBD's four labels \tok{target\_1} to \tok{target\_4} (the customer takes product $k$ during the month), and their mean ROC-AUC is the \emph{official mean}. \emph{Event targets} are five kinds of transaction (type and subtype) made for the first time during the month. \emph{Engineered features} are 114 features built by hand from the schema (counts, amounts, recencies and the profile fields), and \emph{token counts} count the 200 most frequent tokens over time windows with no knowledge of the schema; CatBoost reads both. The \emph{hidden state} is the history decoder's output after an event, and the last one, at the as-of date, is the customer embedding (Section~\ref{sec:embedding}). A \emph{zero-shot score} is the model's probability that the next event is of a given kind, read with no labels and no classification head.

\section{Related work}\label{sec:related}

\begin{figure}[htbp]
\centering
\includegraphics[width=\textwidth]{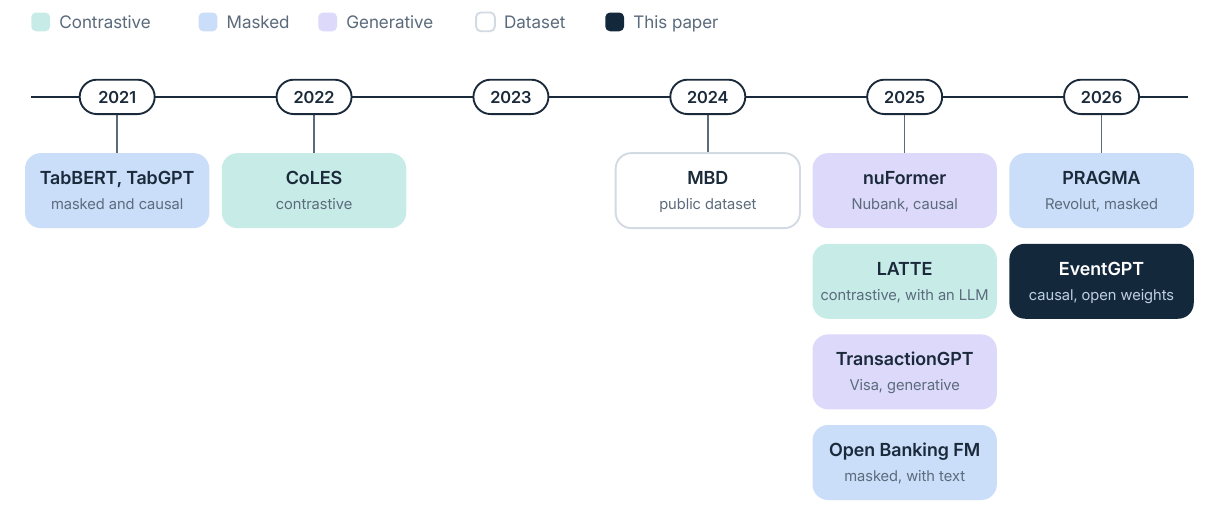}
\caption{Event-sequence models for banking by year of publication. Color marks the pretraining objective; TabBERT and TabGPT are a masked and a causal model from one paper, and MBD is the dataset used here.}
\label{fig:timeline}
\end{figure}

\paragraph{Event-sequence models in banking} Most of the models cited here appeared in 2025 and 2026, with contrastive, masked and generative objectives (Figure~\ref{fig:timeline}). CoLES learns client embeddings by contrastive learning on subsequences of a client's transactions \cite{coles}, and the pytorch-lifestream library implements it with other pretraining methods for event sequences \cite{lifestream}. Its GPT-style objective predicts each field of the next transaction with a separate head; EventGPT instead decodes the next event \emph{token by token}, which gives a joint distribution over its fields. Event Stream GPT, a library for generative transformers over continuous-time medical event streams, also models the dependencies between the fields of an event \cite{esgpt}. LATTE aligns event embeddings with the embeddings a frozen language model gives to short text summaries of a client's behavior \cite{latte}. TabBERT and TabGPT treat each transaction as a row of fields, encode fields into a row embedding and model the rows with a masked or a causal transformer \cite{tabbert}; their code and a synthetic card-transaction dataset are public, without pretrained checkpoints. The MBD paper evaluates these methods, aggregate features and a supervised recurrent network on the same product-propensity task used here \cite{mbd}.

\paragraph{Industrial foundation models} Beyond the designs described in Section~\ref{sec:intro}, nuFormer fuses its embedding with a tabular network at fine-tuning time \cite{nuformer}, and PRAGMA serves credit scoring, fraud and lifetime value from one backbone \cite{pragma}. NVIDIA describes the same recipe, tokenizer to decoder to embedding to classifier, for card transactions \cite{nvidia}. A masked model of transactions with their text descriptions has been trained on data from thousands of North American institutions \cite{openbanking}. Their checkpoints are not publicly available.

\paragraph{Positions from time} Rotary embeddings rotate queries and keys by an angle proportional to a position \cite{rope}. The rotation does not require the position to be an integer, so EventGPT can feed elapsed time in its place. Qwen2.5-Omni sets rotary positions from absolute time, one step per 40 milliseconds, to align audio and video (TMRoPE) \cite{qwenomni}. The Transformer Hawkes Process encodes the time stamps of events with sinusoids in place of positions \cite{thp}.

\paragraph{Factorized attention} TimeSformer divides the attention of a video transformer into attention within each frame and attention across frames \cite{timesformer}. EventGPT factorizes a customer's history the same way, with attention within each event and then across events.

\section{Model}\label{sec:model}

\begin{figure}[htbp]
\centering
\includegraphics[width=\textwidth]{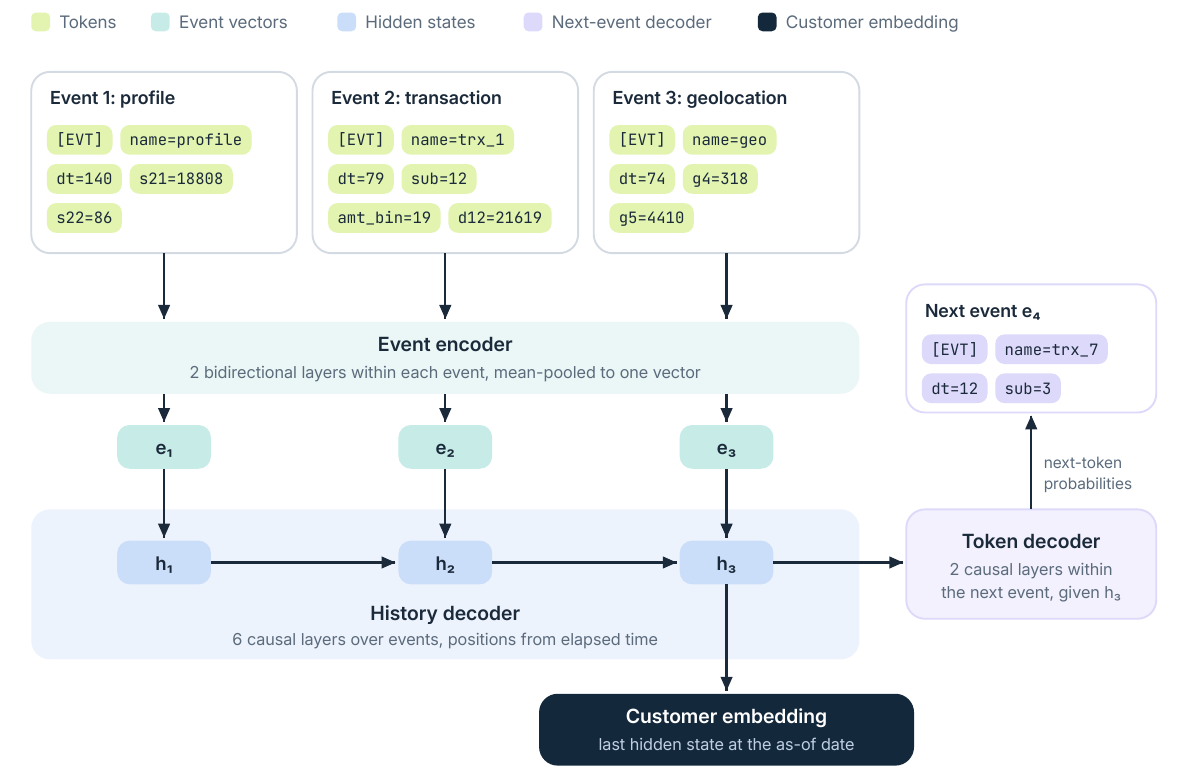}
\caption{The three stages of EventGPT on three MBD events. Token layers read each event on its own and mean-pool it into one vector; causal event layers with temporal RoPE model the event vectors; token layers decode the next event given the hidden state of the previous one.}
\label{fig:arch}
\end{figure}

\begin{figure}[htbp]
\centering
\includegraphics[width=\textwidth]{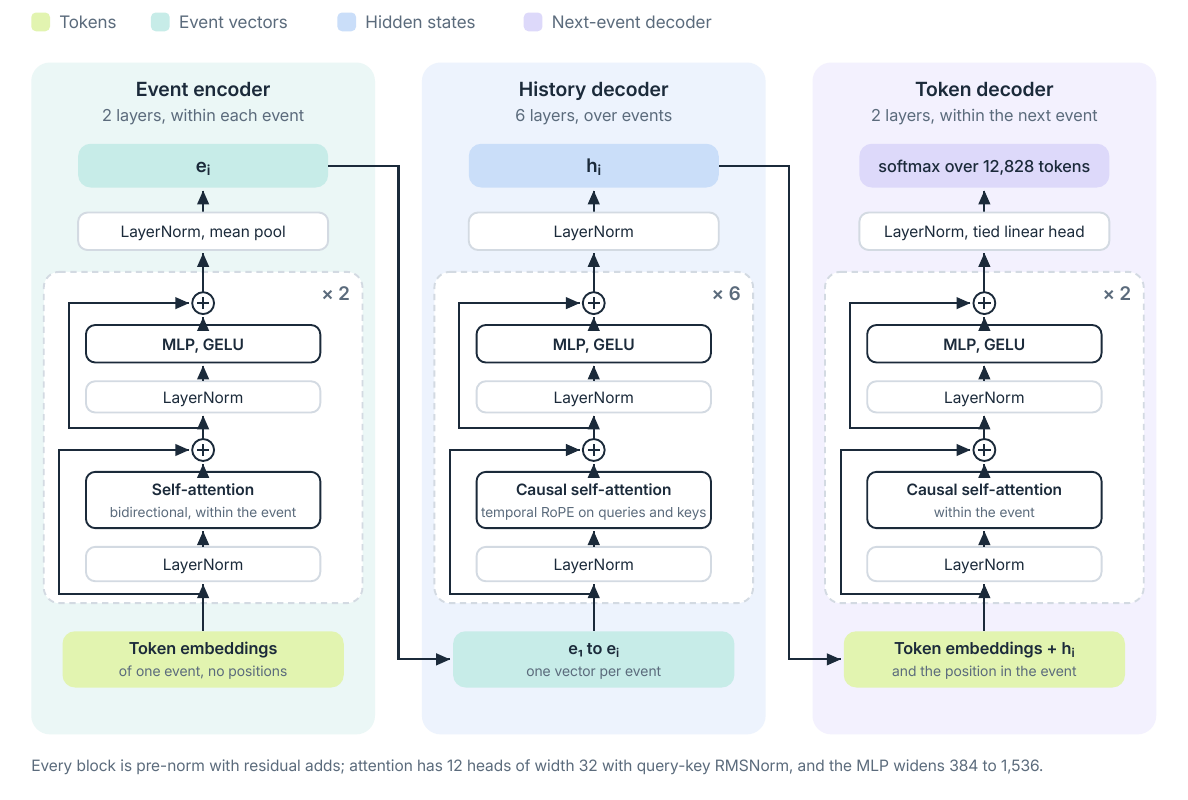}
\caption{The layers inside each stage. The event encoder attends within one event, without positions, and mean-pools the event's tokens into $e_i$. The history decoder attends causally over the event vectors, with temporal RoPE on queries and keys, and gives the hidden state $h_i$. The token decoder adds $h_i$ to every token embedding of event $i+1$ and attends causally within that event; the output layer shares its weights with the token embedding.}
\label{fig:blocks}
\end{figure}

EventGPT has three stages (Figure~\ref{fig:arch}):
\begin{enumerate}
    \item \textbf{Event encoder:} turns each event into a single vector.
    \item \textbf{History decoder:} processes these event vectors for each client.
    \item \textbf{Token decoder:} predicts the tokens of the next event.
\end{enumerate}
Next-token cross-entropy trains all three stages jointly. Each stage consists of a stack of pre-norm transformer blocks; they differ in the inputs over which their attention layers operate (Figure~\ref{fig:blocks}).

\begin{figure}[htbp]
\centering
\includegraphics[width=\textwidth]{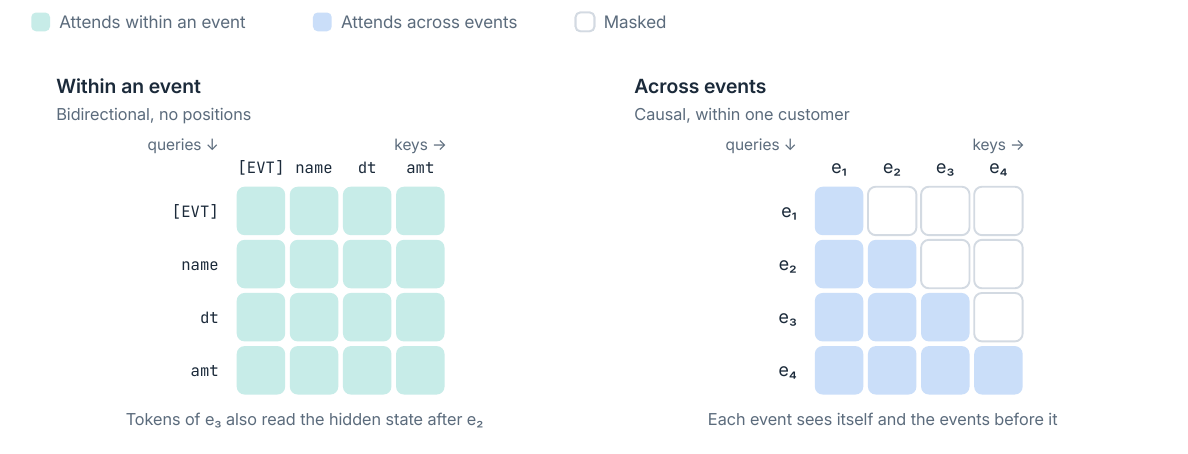}
\caption{Attention at the two levels. Within an event every token attends to every other; across events attention is causal and confined to one customer. The tokens of an event reach earlier events only through the hidden state of the previous event.}
\label{fig:masks}
\end{figure}

\subsection{Event encoder}

An event is a short run of tokens that starts with \tok{[EVT]} (event token), followed by its name, its time gap and its fields as \tok{key=value} tokens (Section~\ref{sec:data}). The fields of an event form a record, whose meaning does not depend on the order in which its fields appear. The event encoder is therefore \emph{bidirectional} within an event and cannot attend across an event boundary. Two layers produce token states that are mean-pooled into one vector per event.

\subsection{History decoder and temporal RoPE}\label{sec:time}

\begin{figure}[htbp]
\centering
\includegraphics[width=\textwidth]{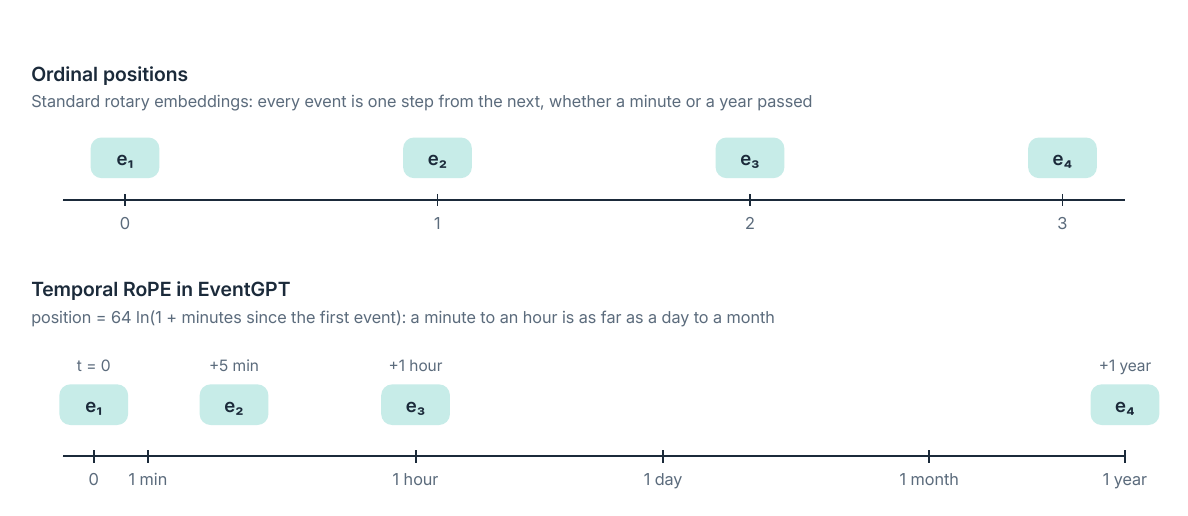}
\caption{Ordinal positions against temporal RoPE, Eq.~(\ref{eq:pos}). With ordinal positions every
event is one step from the next; with temporal RoPE the distance between two events is set by the
ratio of their elapsed times.}
\label{fig:time}
\end{figure}

The history decoder needs each event's vector position, which EventGPT derives from elapsed time. Six causal layers run over each client's event vectors. The hidden state after event $i$ reads the vectors of events $1, \dots, i$, one per event, instead of their tokens. Positions come from elapsed time. TMRoPE is linear in time \cite{qwenomni}, and our temporal RoPE takes its natural logarithm. With $\Delta t_i$ the seconds from the client's first event in the window to event $i$, the rotary position is
\begin{equation}\label{eq:pos}
p_i = 64 \ln\!\left(1 + \frac{\Delta t_i}{60}\right).
\end{equation}
Rotary attention depends on the difference between two positions, and here that difference is $64\ln$ of a ratio of elapsed times. With the 60 seconds added, a gap from a minute to an hour (a ratio of 30.5) is then about as long as a gap from a day to a month (30), and two events a year apart stay far apart even when no event lies between them (Figure~\ref{fig:time}). Elapsed times are derived from the \tok{dt} tokens: each time-gap bucket is represented by its midpoint in seconds. The same gap spans less rotation later in the window: one day corresponds to 44 units between the first and second day, but only 0.09 units between days 699 and 700.

\subsection{Token decoder and the training signal}

\begin{figure}[htbp]
\centering
\includegraphics[width=\textwidth]{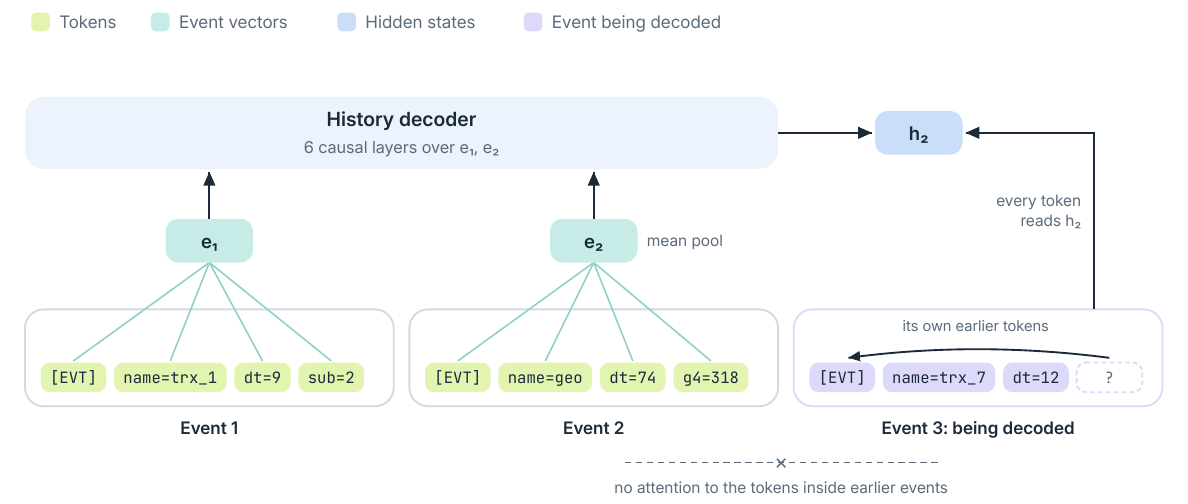}
\caption{What the token decoder reads. Each token of event 3 attends to the earlier tokens of its own event and to $h_2$, the history decoder's hidden state after event 2. It has no path to the tokens inside earlier events, which reach it only through the event vectors and $h_2$.}
\label{fig:decoder}
\end{figure}

Each token of event $i$ receives its embedding, a learned position within the event, and $h_{i-1}$, the history decoder's hidden state after the previous event. For the first event in a window, this state is set to zero. Two causal layers process the tokens and predict the next token through an output layer tied to the input embedding (Figure~\ref{fig:decoder}). The token decoder uses $h_{i-1}$ rather than the pooled representation of event $i$. The event encoder has no separate loss: it receives gradients through the history decoder from the next event's token-prediction loss, so its representation is shaped by how useful the event's fields are for predicting subsequent events.

\subsection{Customer embedding}\label{sec:embedding}

The embedding of a client at an as-of date is read from the history decoder's hidden states for events preceding that date. The last hidden state (\emph{last}) is the embedding used throughout the paper. We compare it with the mean of the hidden states (\emph{mean}) and a recency-weighted mean, which assigns weight $\exp(-a/32)$ to the event $a$ events before the last (\emph{recency}), in Section~\ref{sec:ablations}. All three representations are obtained in one forward pass.

Each history ends with a \emph{now} event, \tok{[EVT] dt=<gap>}, whose gap runs from the last event to the as-of date. The final hidden state therefore corresponds to the as-of date: two clients with identical event histories but different periods of inactivity receive different embeddings.

When a history exceeds the window length, the window retains the most recent events and preserves \tok{[USR]} and the profile event at its beginning. This prevents the static attributes from being discarded for clients with the longest histories.

\subsection{Static attributes}\label{sec:static}

The profile event passes through the same event encoder as all other events, but its four attributes also have a direct path to the classification head. We implement this in two ways. In \emph{late fusion} we append the input embeddings of the four field tokens to the hidden state and fit the classification head on both without retraining the backbone. In \emph{early fusion} we add the attributes during pretraining. We concatenate the four embeddings, project them to the model width with a linear layer initialized at zero and add the result to the vector of every event from the profile event on. We add no profile vector to the \tok{[USR]} event, because its hidden state predicts the profile event's tokens. We take the field tokens from the client's stream and use no knowledge of what the attributes mean.

\paragraph{Training windows} A window drawn at a random offset typically starts partway through a client's history and therefore omits \tok{[USR]} and the profile event that begin every evaluation history. We also train on \emph{aligned} windows, which retain these two events at the beginning (Appendix~\ref{sec:windows}).

\begin{figure}[htbp]
\centering
\includegraphics[width=\textwidth]{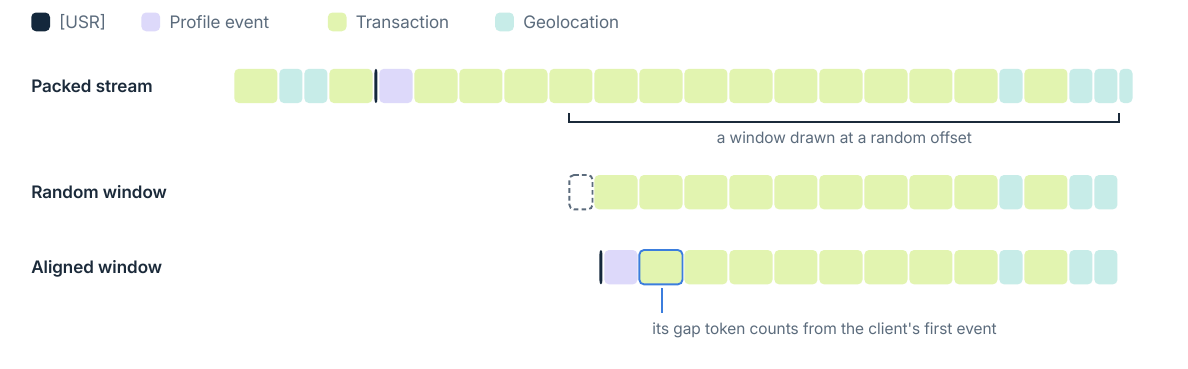}
\caption{A training window drawn at a random offset and the aligned window over the same events, on a client from fold 1, shortened to 110 tokens. The random window starts inside an event and holds no \tok{[USR]} or profile event; the aligned window puts both back in front, and the gap token of its first event counts from the client's first event.}
\label{fig:windows}
\end{figure}

\section{Embeddings as point-in-time features}\label{sec:pit}

An embedding computed once per client and as-of date can serve multiple downstream tasks through a small classification head, with the backbone frozen. In this paper, the same set of embeddings serves ten labels (four product targets, five event targets and churn described in Section~\ref{sec:protocol}), each using logistic regression. A new task then requires only fitting a head on the stored vectors. Revolut and Nubank describe the same arrangement for their backbones \cite{pragma, nuformer}.

This reuse is safe only if the embedding at an as-of date depends exclusively on events before that date. An embedding that incorporates future events inflates the offline results of every task built on it, because all task-specific heads share it. EventGPT enforces this point-in-time property through its architecture and data construction (Appendix~\ref{app:pit}).

This guarantee does not, however, extend to the pretraining population. Pretraining uses events from 2021 and 2022 for the pretraining clients, including events from the test clients' label months. An out-of-time run measures the effect of this overlap in Section~\ref{sec:limits}.

\section{Data}\label{sec:data}

\subsection{MBD-mini}

MBD is a public dataset of bank clients for research on event sequences \cite{mbd}. It holds each client's transactions, geolocation pings and support dialogues, and monthly labels for the purchase of four products. MBD-mini, a 10\% client sample published under CC BY 4.0 \cite{mbdmini}, holds 100,224 clients in five folds of equal size (Table~\ref{tab:mbd}). All fields are anonymized integer codes and amounts are rescaled. Timestamps are in UTC, and the bank's day starts at 21:00 UTC (midnight in Moscow).

\begin{table}[htbp]
\centering
\caption{MBD-mini modalities. Events per client are the 50th, 90th and 99th percentiles.}
\label{tab:mbd}
\begin{tabular}{lrrlr}
\toprule
modality & events & clients & span & per client (p50 / p90 / p99) \\
\midrule
transactions & 38.8M & 98,721 & Jan 2021 -- Dec 2022 & 170 / 1,007 / 2,243 \\
geolocation pings & 66.3M & 72,573 & Dec 2021 -- Dec 2022 & 507 / 2,293 / 5,580 \\
dialogues & 0.22M & 46,320 & Feb 2021 -- Dec 2022 & 3 / 10 / 27 \\
\bottomrule
\end{tabular}
\end{table}

\subsection{Events and tokens}

Each client's stream starts with \tok{[USR]} and a profile event, then interleaves transactions and
geolocation in time order:
\begin{center}
\footnotesize
\setlength{\tabcolsep}{4pt}
\begin{tabular}{ll}
profile & \tok{[EVT] name=profile dt=140 s21 s22 s31 s32} \\
transaction & \tok{[EVT] name=trx\_<type> dt=<b> sub=<subtype> amt\_bin=<b> [cur] s11 s12 d11 d12} \\
geolocation & \tok{[EVT] name=geo dt=<b> g4=<cell> g5=<cell>} \\
\end{tabular}
\end{center}

The profile event states the client's static fields once, and one geolocation event summarizes each day (Appendix~\ref{app:tokens}).

\paragraph{Time and amounts} The gap since the client's previous event becomes a token
\tok{dt=}$\lfloor 8 \ln(1 + \Delta s/8) \rfloor$, capped at 140, so 141 buckets span seconds to
decades. Amounts become one of 21 bins with quantile edges fitted on the pretraining folds. A currency token \tok{cur} appears only when the currency is not the one shared by 99.9\% of transactions.

The stream holds 53.4 million events in all, about 8 tokens per event. Folds 1--4 give the pretraining stream, with 2\% of their clients held out for validation loss: 332 million training tokens and 6.8 million validation tokens. Fold 0 is never seen in pretraining or vocabulary fitting.

\section{Evaluation protocol}\label{sec:protocol}

\subsection{As-of dates and labels}

A row is a client and a month from February 2022 to January 2023, 12 rows per client. As in MBD, the history for a month is every event strictly before the first day of that month \cite{mbd}, here in Moscow time; that day is the month's \emph{as-of date}.

\paragraph{Official targets} \tok{target\_1} to \tok{target\_4} indicate that the client takes product $k$ during the month. They are rare. On the test fold, \tok{target\_1} has 1,072 positives in 243,936 rows (0.44\%), \tok{target\_2} 132 (0.05\%), \tok{target\_3} 934 (0.38\%) and \tok{target\_4} 634 (0.26\%). Of these rows, 1,332 belong to 111 clients with no transactions or geolocation, and so no stream. The classification heads and the baselines both drop them, which leaves 242,604 scored rows. No single event type marks a purchase: the type most often seen for the first time in the month of a purchase covers at most 10\% of its positives.

\paragraph{Event targets} Purchases leave traces in the stream. We rank every transaction kind (type/subtype) on the four training folds by the ratio of its first-occurrence rate in positive client-months to its first-occurrence rate in negative client-months. For \tok{target\_2} the kinds 48/31, 23/49, 26/28 and 48/10 rank first, third, fifth and sixth (rate ratios 215, 94, 66 and 54), and for \tok{target\_3} 50/19 ranks first (25). The kinds ranked second and fourth for \tok{target\_2}, 12/20 and 48/49 (106 and 92), are not used. We chose the five kinds before this ranking and did not keep the script, so we cannot establish which data the choice used. The ranking shows that the selected kinds are strongly enriched in the training folds. It does not reproduce the choice, since two unselected kinds rank above two selected ones. Each of these kinds defines an event target: the kind appears in the client's history for the first time during the month. A \emph{churn} target marks a month with no transaction from a client who had at least one before the as-of date. Both are read from the stream.

\paragraph{The last month} The transaction tables end on 31 December 2022. In January 2023 no client has a transaction, so 99.1\% of that month's rows are labeled churn, against 12.7\% to 20.4\% in the other months, and the month holds 20,032 of the test fold's 55,920 churn positives. A model that can tell the month of the as-of date scores those rows without reading any behavior, and the time since a client's first event gives the month away. With January 2023 included, EventGPT on aligned windows (Section~\ref{sec:windows}) reached 0.910 on churn against 0.830 on random windows; without it, both reach 0.93. We therefore leave January 2023 out of the churn label, in training and in testing, which keeps 35,888 test positives. The event targets have no positives in that month and change by at most 0.014 without it. The official targets are the dataset's own labels and keep all twelve months.

The product targets are rare and drift (Figure~\ref{fig:labels}): \tok{target\_1} falls from 0.56\% of rows in February 2022 to 0.16\% in January 2023, while \tok{target\_2} stays near 0.05\%. The event targets lie between 0.01\% and 0.4\%, and churn rises from 12.7\% to 20.1\% before the last month.

\subsection{Heads and metrics}\label{sec:metrics}

For each label, we fit a logistic regression ($C = 0.1$) on the frozen, standardized embeddings from folds 1--4 and score all rows in fold 0. Positives are rare, so the training sample includes every row positive for at least one label, plus 60,000 rows sampled uniformly from the remaining rows. We reweight the sampled rows to account for their share of the full training set. We report ROC-AUC and PR-AUC (average precision). PR-AUC matters because a ranking can have a high ROC-AUC while still placing mostly negatives near the top. A random ranking has PR-AUC equal to the label prevalence.

We estimate standard errors using DeLong's method \cite{delong}, which expresses ROC-AUC as a mean of placement values, and cluster the variance by client \cite{obuchowski}. The twelve monthly rows for each test client are not independent. We therefore sum each row's influence on the AUC within its client and take the variance across clients:
\begin{equation}
\label{eq:clustered-auc}
\psi_{ij} = \frac{V_{ij} - \widehat{\mathrm{AUC}}}{n_{y_{ij}}},
\qquad
S_i = \sum_{j} \psi_{ij},
\qquad
\widehat{\mathrm{Var}}(\widehat{\mathrm{AUC}}) = \frac{N}{N-1} \sum_{i=1}^{N} S_i^2,
\end{equation}
where $V_{ij}$ is the placement value of row $j$ of client $i$ (for a positive, the share of negatives it outranks; for a negative, the share of positives that outrank it), $n_{y_{ij}}$ is the number of rows in its class, and $N = 20{,}217$.
On simulated data with twelve correlated rows per client, this gives a standard error close to that from a bootstrap over clients (0.0129 versus 0.0132); the unclustered variance gives 0.0102. The same client-level sums give the covariance between two AUCs evaluated on the same rows, allowing us to calculate standard errors for both the official mean and the difference between two models. Since both models rank the same clients, the comparison is paired. Its standard error (about 0.005 on the official mean) is therefore smaller than that of either model's AUC (about 0.006).

\subsection{Zero-shot event targets}\label{sec:zs-protocol}

We score event targets directly with the pretrained model, without a classification head or labeled examples. A target is positive if transaction kind $(X,Y)$ appears in the client's history for the first time during the month. After the history at an as-of date we append the start of a next event, \tok{[EVT] name=trx\_X dt=<g>}, where $g$ is the bucketed gap from the last event to the as-of date. The token decoder gives $P(\tok{name=trx\_X})$ at \tok{[EVT]} and $P(\tok{sub=Y} \mid \tok{name=trx\_X}, \tok{dt=g})$ at the \tok{dt} token, and the score is their product, the probability that the next event is of kind $(X, Y)$. It ranks clients but is not the probability of the label, which counts any occurrence in the month; sampled months give that probability (Section~\ref{sec:zero-shot}). A client who already had the kind before the as-of date cannot be positive under the label's definition, so that client's score is set to zero. This rule uses no labels, and we apply it to the supervised scores as well when comparing the two. For EventGPT the history stages run once per row and the token decoder once per type $X$; the flat decoder is scored the same way and runs whole once per type.

\begin{figure}[htbp]
\centering
\includegraphics[width=\textwidth]{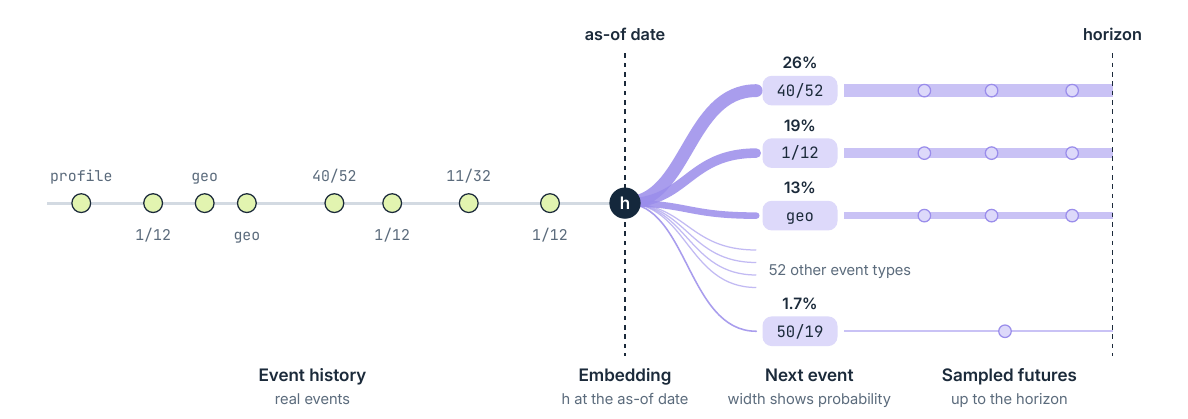}
\caption{Inference from one history. The hidden state $h$ at the as-of date is the embedding read by a supervised head. From $h$ the model gives the distribution of the next event, which the one-step score reads for the target type; the widths are the probabilities for the client of Figure~\ref{fig:futures}, whose next event was a 50/19. Sampling events one after another continues each branch to a horizon, here the end of the label month; 16 sampled futures give the monthly probability of the target.}
\label{fig:inference}
\end{figure}

We read a history in three ways (Figure~\ref{fig:inference}): the hidden state at the as-of date for a supervised head, this one-step score, and months sampled from the model (Section~\ref{sec:zero-shot}).

Zero-shot scoring is evaluated on every test row positive for at least one of the five event targets, together with 20,000 rows sampled uniformly from the test fold. For each target, ROC-AUC is calculated using its own positives and the uniform sample; positives for other targets are not treated as negatives. Uniform sampling of negatives does not bias ROC-AUC. The supervised comparison uses the logistic head on the last hidden state, scored on the same rows.

\subsection{Baselines}

\paragraph{Engineered features} On the same rows we compute 114 features from events before the as-of date. They cover transaction counts and the sum, mean and maximum amount, the count and amount for each of the 15 most frequent event types, and geolocation pings and active days. Each is computed over all history, the last month and the last three months, and the days since the first and the last transaction and the four profile fields (as categorical features) complete the table. CatBoost \cite{catboost} (depth 6, learning rate 0.05, at most 2,000 trees with early stopping on 10\% of training clients) is fitted for each label on the same training rows.

Refitted without one group of features at a time, the official mean falls by 0.010 without the profile fields (0.036 on \tok{target\_4}), by 0.009 without the two timing features, by 0.005 without the full-history window and by 0.001 without geolocation, and it does not change without the amounts. Building these features takes knowledge of the schema.

\paragraph{Token counts} This baseline does not require knowledge of the event schema. For each of the 200 most frequent tokens of the stream (any \tok{key=value}, amount bin or time bucket, together 90.9\% of all non-special tokens), it counts occurrences over all history, the last 91 days and the last 30 days. It also includes the number of events in each window and the number of days since the first and last events, giving 605 features in total. We compute the features either over the full history or over the same 2,048-token window used by EventGPT, and fit CatBoost with the same settings. The baseline uses only token counts and timestamps; it does not require manually designed features for particular event types.

\paragraph{Flat decoder} A GPT-style decoder over the same token stream, with width 384, ten layers and learned positions: 23.5 million parameters against EventGPT's 22.9 million, trained the same way. Its embeddings are the hidden state of the last token, the mean of the token hidden states and a recency-weighted mean over about 256 tokens, which 32 events hold on average.

\paragraph{Published results} The MBD paper reports mean ROC-AUC on full MBD, a population ten times larger with its own five-fold split (Table~\ref{tab:published}). These numbers give context only: MBD-mini is a different sample, and the comparison on full MBD is planned (Section~\ref{sec:limits}). We therefore train CoLES \cite{coles} on our pretraining folds with the MBD paper's transaction configuration (a GRU of width 256 over per-field embeddings, contrastive loss on random slices of a client's transactions, 15 epochs) and read it with the same logistic head as EventGPT. With two seeds it scores 0.771 and 0.772 on the official mean, close to its published 0.773, against 0.786 for EventGPT on aligned windows (seed 42; paired differences 0.015 and 0.014, standard errors 0.005 and 0.006) and 0.816 for the engineered features. CoLES is below EventGPT on \tok{target\_1} to \tok{target\_3}, on every event target and on churn, and above it on \tok{target\_4} (0.784 and 0.795 against 0.754).

\begin{table}[htbp]
\centering
\caption{Mean ROC-AUC over the four product targets on full MBD, as reported by \cite{mbd} (late fusion for the two-modality column).}
\label{tab:published}
\begin{tabular}{lcc}
\toprule
method & transactions & transactions + geo \\
\midrule
aggregation & 0.783 & \\
CoLES & 0.773 & 0.775 \\
TabBERT & 0.762 & 0.764 \\
TabGPT & 0.802 & 0.800 \\
supervised RNN & 0.819 & 0.819 \\
\bottomrule
\end{tabular}
\end{table}

\subsection{Pretraining}

EventGPT has width 384, 12 heads of width 32, two event-encoder layers, six history layers and two token-decoder layers: 22.9 million parameters, of which 4.9 million are the tied token embedding. It is trained with AdamW (learning rate $6 \times 10^{-4}$, $\beta = (0.9, 0.95)$, weight decay 0.1), with linear warmup for 200 steps and cosine decay to 10\% of the peak. Each step processes 16 windows of 2,048 tokens (32,768 tokens) drawn at uniform offsets from the packed stream, for 20,000 steps, about two passes over the data. Windows start mid-event and mid-client, and \tok{[USR]} tokens keep attention within each client.

On one NVIDIA L4 in bfloat16 a run of 655 million tokens takes 1.5 hours (about 118,000 tokens per second), against 3.9 hours (about 47,000) for the flat decoder. Model results are the mean of two seeds, 42 and 43, unless stated otherwise.

The final configuration, labeled \emph{aligned} in the tables, trains on aligned windows (Section~\ref{sec:windows}) and normalizes queries and keys in every attention head by their root mean square (query-key normalization) \cite{qknorm}. Without the normalization, the seed-43 run on aligned windows diverged at step 4,400, at the peak learning rate: its validation loss rose from 1.16 to 1.5 and did not recover. With it, both seeds train smoothly to a validation loss of 1.020 and 1.019, at about 90,000 tokens per second (2.0 hours). The comparisons with the flat decoder use windows at random offsets for both models. The run with early fusion (Section~\ref{sec:static}) adds it to this configuration, at a cost of 0.6 million parameters.

\section{Results}\label{sec:results}

\subsection{Event targets without labels}\label{sec:zero-shot}

Scored from the next-event distribution with no labels (Section~\ref{sec:zs-protocol}), the five event targets reach ROC-AUC 0.79 to 0.92 on both seeds (Table~\ref{tab:zero-shot}). A supervised head on the same model scores 0.045 to 0.061 higher on 26/28 and 48/10 ($p < 0.001$ in every run). On the other three targets the two are level after a Bonferroni correction over 20 paired tests, except 23/49 with aligned windows and seed 43, where the head is higher (Appendix~\ref{app:tests}). The score scaled by recent activity, a form we chose after these scores were known (Section~\ref{sec:ablations}), is higher than the head on 48/31 and lower on 26/28 and 48/10 (Appendix~\ref{app:tests}).

The flat decoder, scored the same way, reaches 0.78 to 0.92, the same range as EventGPT, so zero-shot scoring needs only a model trained to predict the next event.

We also evaluate two simple baselines based on event statistics, without a trained model or labels (Table~\ref{tab:next-event}). A first-order Markov chain over event kinds, fitted on the event streams of the four pretraining folds, estimates the probability that the next event is of the target kind. Using only the last event gives scores of 0.58 to 0.67; averaging the probabilities over the last ten events raises them to 0.72 to 0.84. The number of transactions in the 30 days before the as-of date scores 0.76 to 0.85 on its own. Multiplying the averaged Markov probability by the number of events in those 30 days gives an estimate of the expected count of the target kind at the client's recent activity rate. This combined baseline scores 0.79 to 0.88. EventGPT scores higher on 23/49 and 50/19, by 0.057 and 0.095, respectively. On the other three targets, the differences are within two standard errors. For 48/31, 26/28 and 48/10, recent transaction activity accounts for much of the predictive performance of the zero-shot score.

We chose the five targets by hand (Section~\ref{sec:protocol}), so we also score every transaction kind with at least 50 first occurrences in the test rows, 74 kinds, with the same rule. EventGPT zero-shot on aligned windows has a median ROC-AUC of 0.825 (quartiles 0.772 and 0.894, seed 43; median 0.828 on seed 42), so the five targets are typical of the kinds. Against Markov $\times$ events on the same rows, EventGPT (seed 43) is higher by more than two paired standard errors on 48 kinds, lower on 7 and level on 19. The median difference is 0.042, and 0.057 weighted by positives. The five targets show the model's advantage less than the kinds as a whole.

\begin{table}[htbp]
\centering
\caption{ROC-AUC on the five event targets from event statistics without labels, on the rows of Table~\ref{tab:zero-shot} with the rule for earlier occurrences. Markov: the transition probability to the target kind from the last event, and its mean over the last ten events, fitted on the pretraining folds. Activity: transactions in the 30 days before the as-of date. Markov $\times$ events: the mean transition probability times the number of events in those 30 days. EventGPT is zero-shot on aligned windows, seed 42. The last column is EventGPT minus Markov $\times$ events, with the paired standard error clustered by client.}
\label{tab:next-event}
\footnotesize
\setlength{\tabcolsep}{3pt}
\begin{tabular}{lrcccccc}
\toprule
 & & \multicolumn{2}{c}{Markov} & & Markov & & EventGPT $-$ \\
target & pos. & last event & last ten & activity & $\times$ events & EventGPT & Markov $\times$ events \\
\midrule
first 48/31 & 59 & 0.671 & 0.835 & 0.848 & 0.882 & 0.906 (0.022) & $+0.024$ (0.025) \\
first 23/49 & 195 & 0.667 & 0.833 & 0.811 & 0.860 & 0.916 (0.011) & $+0.057$ (0.016) \\
first 26/28 & 252 & 0.577 & 0.743 & 0.787 & 0.833 & 0.823 (0.013) & $-0.010$ (0.014) \\
first 48/10 & 290 & 0.608 & 0.718 & 0.759 & 0.791 & 0.777 (0.014) & $-0.013$ (0.013) \\
first 50/19 & 770 & 0.667 & 0.802 & 0.773 & 0.811 & 0.906 (0.006) & $+0.095$ (0.007) \\
\bottomrule
\end{tabular}
\end{table}

\begin{table}[htbp]
\centering
\caption{ROC-AUC on the five event targets. Zero-shot scores use no labels and set the score of clients who already had the kind to zero (Section~\ref{sec:zs-protocol}); the flat decoder column is the mean of two seeds, as is the aligned column. The supervised head is the logistic head on the seed-42 embedding (random windows), on the same rows. In brackets the standard error, clustered by client; the seed-43 and aligned columns have standard errors within 0.005 of seed 42. The difference column pairs the seed-42 zero-shot score with the head. Its $p$ comes from the standard error of the difference, which is smaller than either score's own because both scores rank the same clients (Section~\ref{sec:metrics}). The engineered-feature column is the full test fold, so it is comparable in expectation but not row for row. The supervised head and the engineered features are scored without the rule for earlier occurrences.}
\label{tab:zero-shot}
\footnotesize
\setlength{\tabcolsep}{2.5pt}
\begin{tabular}{lrccccccc}
\toprule
 & & \multicolumn{3}{c}{EventGPT zero-shot} & flat decoder & supervised & zero-shot & engineered \\
target & pos. & seed 42 & seed 43 & aligned & zero-shot & head & $-$ head ($p$) & features \\
\midrule
first 48/31 & 59 & 0.916 (0.019) & 0.902 & 0.913 & 0.909 & 0.891 (0.026) & $+0.025$ (0.19) & 0.920 \\
first 23/49 & 195 & 0.918 (0.011) & 0.918 & 0.909 & 0.905 & 0.930 (0.009) & $-0.011$ (0.16) & 0.920 \\
first 26/28 & 252 & 0.829 (0.014) & 0.827 & 0.826 & 0.799 & 0.875 (0.010) & $-0.046$ ($<0.001$) & 0.868 \\
first 48/10 & 290 & 0.791 (0.014) & 0.784 & 0.780 & 0.784 & 0.836 (0.012) & $-0.045$ ($<0.001$) & 0.835 \\
first 50/19 & 770 & 0.906 (0.006) & 0.904 & 0.905 & 0.915 & 0.896 (0.006) & $+0.010$ (0.07) & 0.937 \\
\bottomrule
\end{tabular}
\end{table}

\begin{figure}[htbp]
\centering
\includegraphics[width=\textwidth]{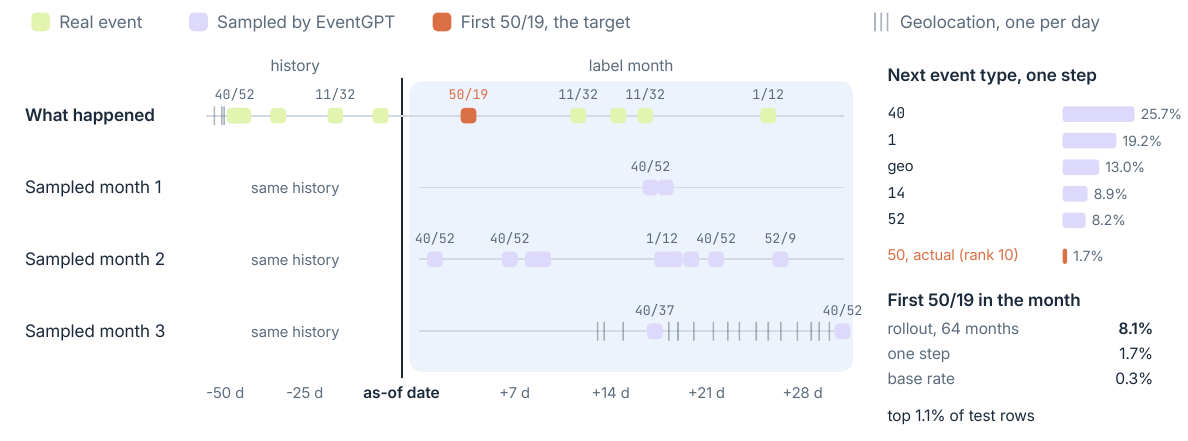}
\caption{One test client at the as-of date 1 December 2022, drawn with a fixed seed among the 770 test rows positive for first 50/19. Top row: the client's last ten events and its label month. Below: three of 64 months sampled from EventGPT (aligned windows, seed 42) from the same history. Right: the model's distribution over the type of the next event, with the type that came next, and the probability of a first 50/19 in the month, from 64 sampled months and from one step, against the share of test rows that are positive.}
\label{fig:futures}
\end{figure}

Figure~\ref{fig:futures} shows what the model expects for one client. Its single-step score for 50/19 ranks in the top 1.1\% of test rows, and months sampled from the model put a first 50/19 in the label month with probability 0.081, against a base rate of 0.003. The client's actual next event was a 50/19, the model's tenth most likely type at 1.7\%.

Without the rule for earlier occurrences the score on 50/19 falls from 0.91 to 0.72. The model predicts the kind for clients who already make it every month. On the rare kind 48/31 the rule makes no difference. The score is the probability of the kind as the next event, while the label covers the whole month.

\paragraph{A month of sampled events}
We also let the model write the label month. For each test row we sample 16 continuations event by event, from the last event to the end of the month, and sum over the sampled steps the probability that the event at the step is of the target kind and falls inside the month, up to the first sampled event of that kind. The mean over continuations estimates the probability that the kind occurs during the month. Averaged over seeds 42 and 43, with every positive and 10,000 uniform test rows, it scores 0.010 to 0.036 above the one-step score on all five targets, by at least three paired standard errors on four of them (Table~\ref{tab:rollout}). The scaled score of Appendix~\ref{sec:ablations}, the one-step score times one plus the client's transactions in the 30 days before the as-of date, was chosen on the test fold. The rollout scores level with it on four targets and 0.012 higher on 50/19, with no such choice. It expects 15.5 transactions per row in the month against 15.5 observed, with a Spearman correlation of 0.81 across rows.

As forecasts of the month itself, the sampled months are weaker. The probability of no transaction in the month scores 0.846 on churn, close to the days since the last transaction alone (0.842) and below the number of transactions in the 30 days before the as-of date (0.884). Matched by kind, a sampled month and the real one share 0.459 of their transactions (counted over the larger of the two in each kind), against 0.545 for the client's previous month repeated and 0.506 for transactions drawn at the client's rate over the last 90 days.

\begin{table}[htbp]
\centering
\caption{Event targets scored from 16 sampled months per row against the one-step zero-shot score and the scaled score, EventGPT on aligned windows, mean of seeds 42 and 43. ROC-AUC on each target's positives and 10,000 uniform test rows, with the rule for earlier occurrences; differences with paired standard errors clustered by client.}
\label{tab:rollout}
\small
\begin{tabular}{lrcccc}
\toprule
target & positives & one-step & scaled & month rollout & rollout $-$ one-step \\
\midrule
first 48/31 & 59 & 0.912 & 0.927 & 0.922 & 0.010 (0.007) \\
first 23/49 & 195 & 0.908 & 0.928 & 0.926 & 0.018 (0.005) \\
first 26/28 & 252 & 0.826 & 0.859 & 0.861 & 0.036 (0.005) \\
first 48/10 & 290 & 0.781 & 0.812 & 0.813 & 0.032 (0.004) \\
first 50/19 & 770 & 0.907 & 0.907 & 0.919 & 0.012 (0.002) \\
\bottomrule
\end{tabular}
\end{table}

\begin{figure}[htbp]
\centering
\includegraphics[width=\textwidth]{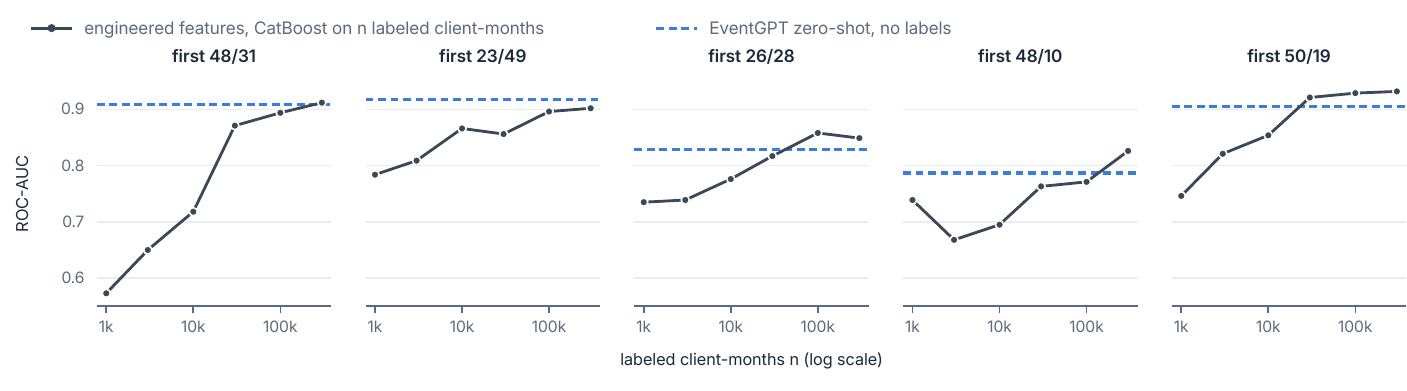}
\caption{ROC-AUC on the test fold of CatBoost on the engineered features, fitted on $n$ client-months drawn at random from the pretraining folds (mean over draws), against EventGPT's zero-shot score (mean of two seeds, no labels). Both apply the rule for earlier occurrences. Values are those of Table~\ref{tab:labels}.}
\label{fig:label-curve}
\end{figure}

\paragraph{Labels needed to match the zero-shot score}
We measure how many labels the engineered features need to match the zero-shot score. We fit CatBoost on the engineered features of $n$ client-months drawn at random from the 699,380 rows of the pretraining folds and score the whole test fold (Figure~\ref{fig:label-curve}, Table~\ref{tab:labels}). A draw without a positive scores 0.5. CatBoost's scores follow the same rule for earlier occurrences.

The engineered features pass the zero-shot score after 10,000 to 30,000 labeled rows on 50/19, the most common kind, and after 30,000 to 300,000 on three others. On 23/49 they are still below it at 300,000 rows and pass it with every positive of the pretraining folds (0.930). This is between about 20 and 400 positives, and more than 260 on 23/49. Draws of 1,000 rows hold at most two positives and draws of 10,000 rows hold 2 to 34, so these row counts are rough thresholds.

Uniform draws spend most labels on negatives. We also draw $n$ positives of the target and 10,000 negatives from the pretraining folds, as a team that labels its known cases and a sample of the rest would (Table~\ref{tab:labels-cc}). The engineered features then pass the zero-shot score with 50 positives on 48/31, 100 on 26/28, 48/10 and 50/19, and 400 on 23/49. Uniform draws pass it at 66 to 394 positives, so either way the zero-shot score is worth about 50 to 400 labeled positives.

\begin{table}[htbp]
\centering
\caption{Engineered features fitted on $n$ positives of the target and 10,000 negatives drawn at random from the pretraining folds, against EventGPT zero-shot (as in Table~\ref{tab:labels}). ROC-AUC on the test fold with the rule for earlier occurrences, mean over five draws; in brackets the standard deviation over draws. The pretraining folds hold fewer than 200 positives of 48/31.}
\label{tab:labels-cc}
\footnotesize
\setlength{\tabcolsep}{3pt}
\resizebox{\textwidth}{!}{%
\begin{tabular}{lcccccccc}
\toprule
 & zero-shot & \multicolumn{7}{c}{engineered features, labeled positives $n$} \\
\cmidrule(lr){3-9}
target & (no labels) & 5 & 10 & 20 & 50 & 100 & 200 & 400 \\
\midrule
first 48/31 & 0.909 & 0.865 (0.025) & 0.833 (0.101) & 0.893 (0.018) & 0.911 (0.009) & 0.914 (0.006) & & \\
first 23/49 & 0.918 & 0.839 (0.023) & 0.857 (0.015) & 0.854 (0.018) & 0.882 (0.018) & 0.897 (0.024) & 0.910 (0.008) & 0.927 (0.002) \\
first 26/28 & 0.828 & 0.800 (0.040) & 0.763 (0.057) & 0.813 (0.039) & 0.826 (0.023) & 0.844 (0.017) & 0.860 (0.011) & 0.874 (0.003) \\
first 48/10 & 0.787 & 0.678 (0.089) & 0.721 (0.030) & 0.746 (0.047) & 0.777 (0.018) & 0.797 (0.026) & 0.817 (0.005) & 0.836 (0.009) \\
first 50/19 & 0.905 & 0.815 (0.031) & 0.847 (0.022) & 0.875 (0.021) & 0.897 (0.007) & 0.921 (0.006) & 0.927 (0.002) & 0.926 (0.010) \\
\bottomrule
\end{tabular}}
\end{table}

\subsection{Frozen embeddings with a supervised head}\label{sec:propensity}

We fit a classification head on the frozen embedding for each label (Table~\ref{tab:main}). Within a 2,048-token window EventGPT scores level with a flat decoder of the same size and with token counts. Both feature tables over the full history score higher.

\begin{table}[htbp]
\centering
\small
\setlength{\tabcolsep}{4pt}
\caption{ROC-AUC on the test fold. Model columns are the mean of two seeds, with a logistic head on the frozen embedding; EventGPT is trained on windows at random offsets or on aligned windows
(Section~\ref{sec:windows}). Token counts and engineered features are fitted with CatBoost. Churn leaves out January 2023 (Section~\ref{sec:protocol}). The engineered features are fitted on the training rows drawn once January 2023 had left the churn label, the other columns on rows drawn before, which also held the 20,000 January churn positives. Every column scores the event targets without the rule for earlier occurrences. PR-AUC for the official targets is in Table~\ref{tab:main-pr}.}
\label{tab:main}
\begin{tabular}{lccccccc}
\toprule
 & engineered & \multicolumn{2}{c}{token counts} & \multicolumn{2}{c}{flat decoder} & \multicolumn{2}{c}{EventGPT, last} \\
target & features & full history & 2,048 tokens & last & recency & random & aligned \\
\midrule
\tok{target\_1} & 0.787 & 0.781 & 0.757 & 0.692 & 0.738 & 0.740 & 0.756 \\
\tok{target\_2} & 0.843 & 0.849 & 0.792 & 0.725 & 0.823 & 0.829 & 0.850 \\
\tok{target\_3} & 0.816 & 0.809 & 0.788 & 0.729 & 0.783 & 0.776 & 0.789 \\
\tok{target\_4} & 0.818 & 0.774 & 0.770 & 0.709 & 0.765 & 0.753 & 0.752 \\
mean & 0.816 & 0.803 & 0.777 & 0.714 & 0.777 & 0.774 & 0.787 \\
\midrule
first 48/31 & 0.920 & 0.905 & 0.913 & 0.794 & 0.909 & 0.889 & 0.900 \\
first 23/49 & 0.920 & 0.905 & 0.915 & 0.796 & 0.918 & 0.918 & 0.929 \\
first 26/28 & 0.868 & 0.879 & 0.874 & 0.745 & 0.844 & 0.860 & 0.868 \\
first 48/10 & 0.835 & 0.857 & 0.846 & 0.715 & 0.793 & 0.804 & 0.807 \\
first 50/19 & 0.937 & 0.935 & 0.922 & 0.769 & 0.856 & 0.837 & 0.849 \\
churn & 0.956 & 0.955 & 0.955 & 0.875 & 0.899 & 0.935 & 0.932 \\
\bottomrule
\end{tabular}
\end{table}

\paragraph{Hierarchy against the flat decoder} We train both models on windows at random offsets. EventGPT scores 0.774 on the official mean and the flat decoder 0.777. The flat decoder's two seeds differ by 0.018, more than this gap (Table~\ref{tab:flat}). The flat decoder reaches 0.777 only by pooling about 256 token states. Its last token scores 0.714 and is below EventGPT's last hidden state on every label. EventGPT also scores higher on churn (0.935 against 0.899). EventGPT trains at 2.5 times the flat decoder's throughput at 2,048 tokens and 4.3 times at 4,096 (Table~\ref{tab:gpu}) but has a higher next-token loss (1.02 against 0.94). We feed its token decoder only the hidden state after the previous event. The loss gap is largest on the geolocation cell, a field that repeats from recent events: top-1 accuracy 0.57 and 0.58 against 0.68 and 0.69. On the event name the two are level (0.57 against 0.58). A lower next-token loss therefore does not give a better embedding.

\paragraph{Against feature tables} Within the same 2,048-token window EventGPT scores level with token counts on random windows (0.774 against 0.777) and above them on aligned windows (0.787). Over the full history token counts reach 0.803 and the engineered features 0.816. Token counts also score higher on churn, within the window and over the full history (0.955 against 0.935 and 0.932). On transactions alone EventGPT is level with token counts over the full transaction history (Section~\ref{sec:ablations}).

\subsection{Aligned training windows}\label{sec:aligned}

Aligned windows raise the official mean from 0.774 to 0.787 on both seeds (Table~\ref{tab:main}). They also raise the classification head on every event target and leave the zero-shot scores unchanged (Table~\ref{tab:zero-shot}). Without query-key normalization aligned windows score 0.788 (seed 42, Table~\ref{tab:window}), so the gain comes from the windows. Churn does not change. The large churn gain that aligned windows first seemed to give came from recognizing the last month (Section~\ref{sec:protocol}). Without January 2023 the gain on the product targets holds (0.787 against 0.774). The mismatch between random training windows and evaluation histories costs the product targets about 0.01.

\subsection{Reducing the gap to engineered features}\label{sec:gap}

We score every model on the same test rows as the engineered features, and fit the heads of both seeds on the same training rows (Table~\ref{tab:gap}). The hidden state scores 0.030 and 0.028 below the engineered features. Late fusion of the profile fields brings it to 0.016 and 0.015 below. With early fusion and full-history token counts the gap falls to 0.006 and 0.009 ($p = 0.16$ and 0.057).

\begin{table}[htbp]
\centering
\caption{The gap to the engineered features, aligned windows with query-key normalization, two pretraining seeds. Official mean ROC-AUC with its difference from the engineered features (0.816), a 95\% interval and $p$ from the paired standard error, clustered by client. The last two rows are rank-averaged ensembles: they average, with equal weights, the ranks given by the embedding with late-fused profile fields and token counts and by CatBoost on token counts over the full history or on the engineered features. The timing and token-count rows each add their features to the hidden state and the late-fused profile fields, not to the row above. Fitted on these training rows, the token counts reach 0.783 against 0.803 in Table~\ref{tab:main}, most of it on \tok{target\_2} (0.780 against 0.849).}
\label{tab:gap}
\footnotesize
\setlength{\tabcolsep}{3pt}
\resizebox{\textwidth}{!}{%
\begin{tabular}{lcccccc}
\toprule
& \multicolumn{3}{c}{seed 42} & \multicolumn{3}{c}{seed 43} \\
\cmidrule(lr){2-4}\cmidrule(lr){5-7}
model & mean & difference [95\% int.] & $p$ & mean & difference [95\% int.] & $p$ \\
\midrule
\multicolumn{7}{l}{\emph{EventGPT}} \\
hidden state & 0.786 & $-0.030$ [$-0.038$, $-0.022$] & $<0.001$ & 0.788 & $-0.028$ [$-0.037$, $-0.020$] & $<0.001$ \\
\quad + profile fields, late fusion & 0.800 & $-0.016$ [$-0.025$, $-0.007$] & $<0.001$ & 0.801 & $-0.015$ [$-0.026$, $-0.005$] & 0.004 \\
\multicolumn{7}{l}{\emph{EventGPT with early fusion of the profile fields}} \\
hidden state & 0.784 & $-0.032$ [$-0.041$, $-0.022$] & $<0.001$ & 0.793 & $-0.023$ [$-0.032$, $-0.015$] & $<0.001$ \\
\quad + profile fields, late fusion & 0.803 & $-0.013$ [$-0.023$, $-0.004$] & 0.007 & 0.807 & $-0.009$ [$-0.018$, $+0.000$] & 0.057 \\
\quad + profile fields + timing & 0.804 & $-0.012$ [$-0.022$, $-0.003$] & 0.009 & 0.810 & $-0.006$ [$-0.014$, $+0.003$] & 0.19 \\
\quad + profile fields + token counts & 0.810 & $-0.006$ [$-0.015$, $+0.003$] & 0.16 & 0.807 & $-0.009$ [$-0.018$, $+0.000$] & 0.057 \\
\midrule
ensemble with token counts & 0.816 & $+0.000$ [$-0.006$, $+0.007$] & 0.90 & 0.816 & $+0.001$ [$-0.006$, $+0.007$] & 0.87 \\
ensemble with engineered features & 0.830 & $+0.014$ [$+0.010$, $+0.019$] & $<0.001$ & 0.830 & $+0.014$ [$+0.009$, $+0.019$] & $<0.001$ \\
\bottomrule
\end{tabular}}
\end{table}

Appendix~\ref{app:gap} breaks the gap down by static attributes, timing, history before the window, knowledge of the schema and amount resolution.

\paragraph{Rank-averaging ensembles} We test the embedding inside the engineered features, as 32 principal components or as a CatBoost embedding feature. The official mean stays the same or falls (Appendix~\ref{sec:finetune}). We then average the ranks of EventGPT with the profile fields and token counts and of the engineered features, with equal weights. The ensemble reaches 0.830 on both seeds, 0.014 above the engineered features, with intervals that exclude zero (Table~\ref{tab:gap}). Any second model can raise a rank average, so we build the same ensemble without EventGPT: a logistic head on the profile fields, one-hot encoded, and the same token counts, rank-averaged with the engineered features. On the ten fold and seed pairs of Table~\ref{tab:folds} it scores 0.001 to 0.008 above the engineered features (mean 0.004). The ensemble with EventGPT scores 0.007 to 0.016 above it (mean 0.012, $p < 0.01$ on each pair, paired standard errors clustered by client). The gain is largest on \tok{target\_2} and \tok{target\_4} and appears on four of the five event targets. Churn does not change. An ensemble with full-history token counts in place of the engineered features reaches 0.816, the level of the engineered features, with no knowledge of the schema.

We fix the weights in advance. With any weight on EventGPT from 0.3 to 0.6 the official mean stays within 0.002 of its peak on both seeds. We score the ensemble and the engineered features on the same test rows with the same fit of the engineered features, so the variation between fits does not enter the 0.014. That variation moves full-history token counts from 0.803 to 0.783 between two draws of training rows. A second draw of the training rows of EventGPT's classification head (seed 42) gives 0.808 against 0.810 and the same ensemble score, 0.830.

\paragraph{Top of the ranking} On PR-AUC and recall among the top-ranked test rows the ensemble is level with or slightly above the engineered features (Appendix~\ref{app:top}).

\paragraph{Five-fold cross-validation} We repeat the configuration of Table~\ref{tab:gap} with each of folds 1--4 as the test fold. We pretrain on the other four folds with two seeds and refit the engineered features and token counts on each split (Table~\ref{tab:folds}). The ensemble scores above the engineered features on all ten fold--seed pairs, by 0.014 to 0.018 ($p < 0.001$ on each). EventGPT with the profile fields and token counts scores 0.002 to 0.009 below the engineered features, and no difference is significant ($p \geq 0.057$). On PR-AUC the ensemble scores 0.021 against 0.019 for the engineered features, averaged over folds, and is level on fold 3. In the top 1\% of test rows the two hold the same share of positives (0.150). EventGPT with the profile fields and token counts holds 0.132.

\begin{table}[htbp]
\centering
\caption{Official mean ROC-AUC with each fold as the test fold, EventGPT with early fusion of the profile fields as in Table~\ref{tab:gap}, mean of two pretraining seeds. Engineered features and token counts are refitted on each split. The last row gives the mean over folds and, in parentheses, the standard deviation. The hidden-state column is the early-fusion model; its fold-0 value (0.789) differs from the aligned column of Table~\ref{tab:main} (0.787), which is trained without early fusion.}
\label{tab:folds}
\small
\setlength{\tabcolsep}{4pt}
\begin{tabular}{lccccc}
\toprule
 & engineered & \multicolumn{2}{c}{EventGPT} & & ensemble $-$ \\
\cmidrule(lr){3-4}
test fold & features & hidden state & + profile, counts & ensemble & engineered \\
\midrule
0 & 0.816 & 0.789 & 0.808 & 0.830 & $+0.014$ \\
1 & 0.804 & 0.790 & 0.802 & 0.819 & $+0.016$ \\
2 & 0.810 & 0.782 & 0.805 & 0.826 & $+0.016$ \\
3 & 0.813 & 0.794 & 0.809 & 0.830 & $+0.017$ \\
4 & 0.798 & 0.780 & 0.796 & 0.814 & $+0.015$ \\
\midrule
mean & 0.808 (0.007) & 0.787 (0.006) & 0.804 (0.005) & 0.824 (0.007) & $+0.016$ (0.001) \\
\bottomrule
\end{tabular}
\end{table}

\subsection{Ablations and length of history}

We change one element of the model or its data at a time (Appendix~\ref{sec:ablations}). Pretraining on transactions only raises the official mean from 0.774 to 0.793. Temporal RoPE scores 0.781 against 0.778 for learned ordinal positions, a difference smaller than the one between seeds. Dropping the loss on gaps, amount bins and geohash cells leaves the official mean unchanged (0.775 against 0.774). Windows of 4,096 and 8,192 tokens do not raise EventGPT's official mean, while token counts gain 0.024 from 2,048 to 8,192 tokens (Appendix~\ref{sec:context}).

\section{Discussion}\label{sec:discussion}

On MBD, next-event pretraining scores five event targets without labels at ROC-AUCs that CatBoost on the engineered features reaches only with labels. On four targets, CatBoost needs 10,000 to 300,000 randomly sampled client-months, and on the fifth it remains below the zero-shot score even with 300,000. Event statistics without a trained model, using a Markov chain over event kinds scaled by recent activity, come within two standard errors of EventGPT on three of the five targets but fall below it on 48 of 74 kinds, with a median gap of 0.042. Product purchases require a classification head, and the frozen embedding then scores 0.021 below the engineered features, averaged over five test folds. Adding profile fields and full-history token counts reduces the gap to 0.004. As a second model alongside the engineered features, EventGPT adds 0.014 to 0.017 on each of the five folds, with a mean gain of 0.016. Of this gain, 0.012 remains against the same ensemble built from profile fields and token counts without EventGPT. In the top 1\% of test rows, the ensemble and engineered features recall the same share of positives (0.150). At the top 0.5\% and 5\%, the ensemble recalls slightly more on average (0.103 against 0.100 and 0.362 against 0.358), although the effect varies by target. Precision is the same at all three cutoffs. The results suggest that next-event pretraining captures behavioral information useful both for direct event scoring and for ranking alongside conventional features. Static attributes and events outside the model's window remain important, however, and are better represented by the engineered features.

\paragraph{Hierarchy benefits} 
The zero-shot results primarily reflect next-event pretraining rather than the hierarchical architecture: the flat decoder also reaches ROC-AUCs of 0.78 to 0.92. With both models trained on random-offset windows, their official means are similar (0.777 for the flat decoder and 0.774 for EventGPT), while the two flat-decoder seeds differ by 0.018. The hierarchy provides a more useful \emph{event-level representation}: its last hidden state is the customer embedding, without an additional pooling choice. The flat decoder's last-token representation scores 0.714; a recency-weighted mean over about 256 token states is needed to reach 0.777.

On churn, EventGPT scores 0.935 against 0.899 for the flat decoder, at both window lengths used to train the flat decoder (Table~\ref{tab:window}). We add four recency features to the churn classification task: the days since the last event and since the last transaction, and the numbers of events and transactions in the 30 days before the cut. These raise the flat decoder to 0.933 on both seeds and EventGPT to 0.940 and 0.941, suggesting that the EventGPT state already encodes much of the recency signal. With learned ordinal positions in place of temporal RoPE, EventGPT scores 0.927 and 0.928 on churn, compared with 0.933 with temporal RoPE at the same learning rate. This suggests that temporal RoPE contributes to the difference, although the comparison does not isolate its effect from other training differences.

The hierarchy also reduces training cost. The flat decoder runs all ten layers over every token in the window, with attention cost growing quadratically in the window length. EventGPT runs four of its ten layers over tokens: two in the event encoder and two in the token decoder. Under block attention, a token attends only to tokens in its own event, about eight. The six history layers operate over events, about one for every eight tokens. Thus, each history layer passes one eighth as many tokens through the MLP and attends over about 1/64 as many token pairs as a flat layer. Counted in layer passes, EventGPT makes about 4.75 passes per token, against 10 for the flat decoder.

On one L4 GPU, 20,000 steps (655 million tokens) take EventGPT about 1.5 hours at every window length from 2,048 to 8,192 tokens, and the flat decoder 3.9 hours at 2,048 tokens and 6.2 hours at 4,096 (Table~\ref{tab:gpu}). We did not train the flat decoder at 8,192 tokens (Appendix~\ref{sec:context}). At inference, a sampled event requires one history-decoder step over the event states, about 250 in a full window, and one token-decoder pass over the event's own tokens. A flat decoder with a KV cache requires about eight steps per event, each operating over all token states in the window.

The next-event distribution also answers queries that a classification head cannot. Appending a hypothetical event to a history and reading the distribution after it gives the model's prediction for clients with that history and event. This prediction is \emph{conditional on the history}: it reflects which clients had such events in MBD and does not estimate the effect of causing one. The log-likelihood of a client's recent events under the model could also flag unusual behavior or a change in that client's activity. MBD has no fraud or distress labels with which to test this use.

The open question is whether the model can carry history beyond its window. We would next test a compressed summary of the earlier history, read as a prefix before the window. Another test would use a lender's mixed applications, repayments and account events, where a feature table is harder to build.

\paragraph{Pretraining per institution}\label{sec:per-institution} Each institution pretrains an event model on its own events. The architecture, the rules that turn an event into tokens and the training procedure carry over to another institution. The weights do not transfer the way a language model's do. A language model reads text through a vocabulary that all its users share, so one checkpoint serves any text in its languages. An event model's vocabulary is the issuer's event schema: its tokens name the event types, field codes and categories of one bank's systems, and its amount bins are quantiles of that bank's amounts.

Every field value is an integer code assigned by the bank. Thus, tokens such as \tok{name=trx\_1} and \tok{s21=18808} have no shared meaning across institutions. The 12,828 tokens and the 21 amount-bin edges were fitted on the pretraining folds of this one population, and a token id has its meaning only together with the vocabulary file it was fitted with. Even within one institution, fitting the vocabulary again assigns ids anew, so we store a fingerprint of the vocabulary with every checkpoint and refuse to score a checkpoint against a vocabulary it was not trained on. The time-gap tokens are the exception, since $\lfloor 8\ln(1 + \Delta s/8) \rfloor$ means the same at every bank.

A checkpoint that transfers would need tokens that mean the same at every bank. The time-gap tokens already do. Event types and fields could be named by text descriptions, as in the masked model trained on data from thousands of North American institutions \cite{openbanking}, or mapped to a shared schema of event kinds. Amounts would need bins on a common scale, such as currency units, in place of one bank's quantiles. Codes with no shared meaning, such as counterparty types, could be hashed into a fixed number of buckets, which fixes the vocabulary size but gives a code no meaning across banks. We have not tested any of these.

Serving such a model starts with pretraining on the institution's own history, and the comparisons in this paper are between methods trained on the same data. Privacy and competition keep the industrial checkpoints of Section~\ref{sec:related} private in any case, and without its issuer's schema another institution could not use one.

\section{Limitations}\label{sec:limits}

The results are on MBD-mini, for which no published numbers exist, so the comparison with the methods of the MBD paper (Table~\ref{tab:published}) awaits a run on full MBD. MBD's events are mostly transactions with one field layout, which gives the event encoder less to do than histories that mix applications, repayments and other events with different fields. The geolocation stream starts a year after the transactions, so early as-of dates carry little of it, and dialogue embeddings are not used. Zero-shot scoring covers only targets that are events in the stream; MBD's product targets are external labels and need a head. Zero-shot scoring predicts the next event, whereas the label records whether the event occurs at any point during the month. Its comparison with the supervised head is therefore evaluated on a sample of the test fold.

The reported standard errors reflect variation across test clients, not variation from training-set sampling or model initialization. We repeat the main comparison on five test folds and every other result on fold 0 alone. On fold 0, the engineered features and the ensemble score 0.008 and 0.006 above their five-fold means. A head fitted on another draw of training rows moves \tok{target\_2}, with 132 test positives, by up to 0.07 and the official mean by up to 0.02, so differences of that size between models fitted on different draws are not established. We ran fine-tuning once, on one seed with one setting of the learning rates, and chose its step on clients from the pretraining folds. We ran widths beyond 384 with one seed, trained no model larger than 140 million parameters and did not tune the regularization of the head for the wider embeddings.

The split is by client, as in MBD. Pretraining reads the pretraining clients' events from 2021 and 2022, which include the test clients' label months, although each test history ends before its label month. We separate the two with an out-of-time run: we pretrain on transactions before February 2022 only, with 54\% of the tokens. We compare it with a transactions-only model pretrained over the full period, which scores 0.793 (mean of two seeds). With the same 20,000 steps, the out-of-time model scores 0.789, with churn scores of 0.950 and 0.949. We use transactions alone because the geolocation stream starts in December 2021. At the same number of steps, the out-of-time run sees each token about 4.4 times instead of 2.4. We therefore also compare runs with matched token exposure: at 11,000 steps, the out-of-time run sees each token about 2.4 times. Its official mean is 0.787 against 0.793 for full-period pretraining, with churn scores of 0.947 and 0.949. Aligned windows also give the model each client's tenure through the gap token of the first event, and with tenure the calendar time of the as-of date; the last-month effect on churn (Section~\ref{sec:protocol}) is one consequence. In production, as-of dates lie beyond the pretraining period, and what the model learned about calendar time may no longer hold.

\section{Code and data availability}\label{app:code}

The code (tokenization, model, training, evaluation and baselines) is at \url{https://github.com/deburky/EventGPT}, and the pretrained checkpoints are on the Hugging Face Hub as \tok{deburky/EventGPT}. MBD-mini is available on the Hugging Face Hub as \tok{ai-lab/MBD-mini} under CC BY 4.0 \cite{mbdmini}.

\appendix
\section{Paired tests of the zero-shot scores}\label{app:tests}

We test each zero-shot score against the supervised head of the same model and seed. The head scores 0.045 to 0.061 higher on 26/28 and 48/10 for both seeds of both models ($p < 0.001$). On 23/49 the head scores 0.008 and 0.011 higher with random windows ($p = 0.33$ and 0.16) and 0.025 and 0.034 higher with aligned windows ($p = 0.003$ and $< 0.001$). On 48/31 the zero-shot score is 0.029 higher with aligned windows and seed 43 ($p = 0.008$). On 50/19 it is 0.012 higher with random windows and seed 43 ($p = 0.04$). The other six runs on these two targets show no significant difference ($p \ge 0.07$). We run 20 paired tests: five targets for both seeds of both models. At a Bonferroni threshold of 0.0025 the head stays significantly higher on 26/28 and 48/10 in every run and on 23/49 with aligned windows and seed 43. Neither zero-shot win passes this threshold. Training on aligned windows leaves the zero-shot scores unchanged.

We also multiply the score by one plus the client's transactions in the 30 days before the as-of date, a form we chose after the scores of Table~\ref{tab:zero-shot} were known (Section~\ref{sec:ablations}). We test the scaled score against the head for the same five targets and four runs. It scores 0.013 to 0.043 higher on 48/31 and 0.012 to 0.029 lower on 26/28 and 48/10. On 23/49 and 50/19 the two differ by at most 0.014 ($p \ge 0.027$). These 20 tests and the 20 above make 40, for a Bonferroni threshold of 0.00125. The head's results above still pass it. Of the new tests only the head on 26/28 with aligned windows and seed 42 passes ($p < 0.001$). The scaled score on 48/31 with aligned windows and seed 43 comes closest ($p = 0.0014$).

\section{Delta scoring}\label{app:delta}

\begin{figure}[htbp]
\centering
\includegraphics[width=\textwidth]{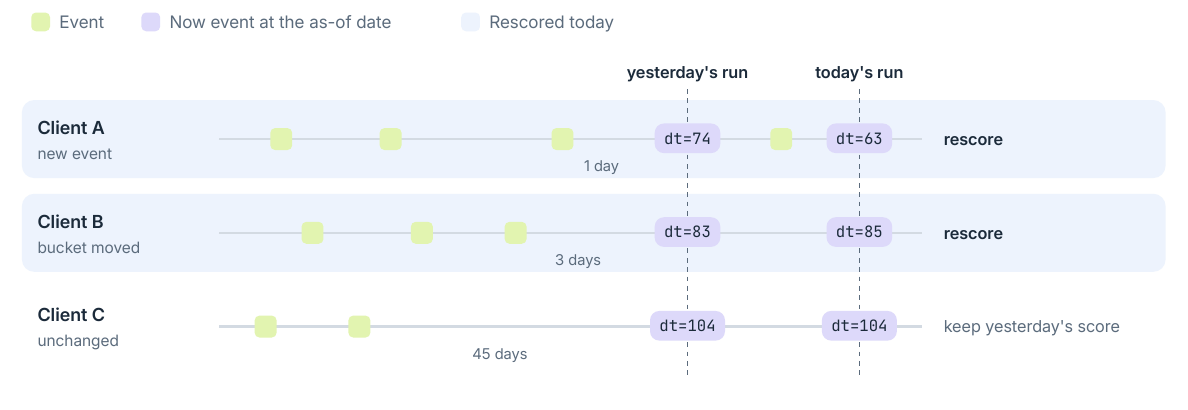}
\caption{Delta scoring with a now event. Client A has a new event since yesterday's run. Client B has none, but its silence grew from three to four days and its now token moved from \tok{dt=83} to \tok{dt=85}. Client C has been silent for 45 days, and one more day leaves its now token at \tok{dt=104}. The job rescores A and B and keeps C's scores from yesterday.}
\label{fig:delta}
\end{figure}

A client's history ending at its as-of date lets a daily job rescore only the clients whose input changed (Figure~\ref{fig:delta}). A client's embedding depends only on the tokens of its history, so a client with no new event keeps its embedding and its scores. The now event is the exception. Its \tok{dt} token measures the silence up to the as-of date and moves to the next bucket as the silence grows. A bucket spans about one eighth of the gap, so a client silent for $g$ days changes bucket on about $8/g$ of days once $g$ exceeds eight. The job therefore rescores every client with a new event and every client whose now token changed. It then reproduces a full pass exactly, provided the streams are built with the vocabulary of the checkpoint (Section~\ref{sec:per-institution}).

We count both groups for each day of 2022 on MBD-mini. With geolocation, 57\% of clients have a new event on an average day and a further 32\% have a new now token, so the job rescores 90\% of clients. On transactions alone the shares are 31\% and 48\%, and the job rescores 79\%. The saving is small on MBD because most of its clients act every few days. It grows with the share of clients who stay silent for weeks.

\section{Fine-tuning and gradient boosting on the embedding}\label{sec:finetune}

We fine-tune the aligned checkpoint of seed 42, trained without query-key normalization, end to end. We add one logistic output per label and train on the training rows of the classification heads. We train only the head for the first 2,000 steps and then every weight for one pass, at a peak learning rate of $3 \times 10^{-5}$ for the pretrained weights and $10^{-3}$ for the head. We choose the step by the official mean on a tenth of the training clients. The official mean stays at 0.788. The five event targets rise in this run, which has one seed and one setting of the learning rates. On 48/31 and 23/49 the fine-tuned model scores above the engineered features (0.940 and 0.931 against 0.920). This run kept January 2023 in the churn label, so its churn score includes the last-month effect. The validation clients belong to the pretraining folds, and their official mean is 0.06 above the test mean.

We test the embedding inside CatBoost, as 384 numeric columns or as one embedding feature. With an embedding feature CatBoost computes its own features, by default linear discriminant projections and counts of nearest training rows per class. The numeric columns score 0.775 and the embedding feature 0.700, against 0.788 for the logistic head. The embedding feature scores 0.501 on \tok{target\_4}. We also add the embedding feature to the engineered features. Their official mean falls from 0.815 to 0.808; these rows were drawn before January 2023 left the churn label, so the engineered features score 0.815 here against 0.816 in Table~\ref{tab:main}.

\section{Training windows and tenure}\label{sec:windows}

Windows drawn at random offsets usually start partway through a client's history and lack \tok{[USR]} and the profile event. Every evaluation history opens with both. We remove this mismatch with \emph{aligned} windows (Figure~\ref{fig:windows}). When a window starts partway through a client's history, we put \tok{[USR]} and the profile event back at its start, as we do when we truncate a long evaluation history. We stamp the profile event at the client's first event. We start both the \tok{dt} token of the next event and the temporal RoPE positions of Eq.~(\ref{eq:pos}) at the client's first event. Apart from the profile event and this one \tok{dt} token, an aligned window holds the same tokens as a random one. We build evaluation and zero-shot histories the same way for these models.

\section{Point-in-time guarantees}\label{app:pit}

\paragraph{In the architecture} The history decoder is causal, so the hidden state after event $i$ depends on events $1, \dots, i$ only. The event encoder reads in both directions only within one event, and the token decoder reaches earlier events only through $h_{i-1}$. No token of event $i$ has a path to the encoded vector of event $i$ itself. The embedding after the last event before an as-of date is the same whether or not later events exist in the stream. Evaluation also builds each input to end at its as-of date, so later events are never present in it.

\paragraph{In the data} How events are built matters as well, and each step of Section~\ref{sec:data} keeps the property.
\begin{itemize}
\item The history for an as-of date is every event strictly before the first day of the month, in the bank's own time zone.
\item A daily geolocation event is stamped at the day's last ping, so a summary of the day never appears before the pings it summarizes.
\item The profile event carries fields that never change within a client and is stamped at the client's first event, so it enters a history only when some event does.
\item The vocabulary and the amount-bin edges are fitted on the pretraining folds only, and the test fold is never seen in pretraining.
\end{itemize}

Because the history itself fixes its as-of date, an embedding can be stored in a feature store like any other dated feature. It is keyed by client, stamped with its as-of date instead of the time it was computed, and joined to labels by the store's point-in-time join, which returns for each label the latest embedding before it. The join decides which embedding a label sees, and since each history ends at its as-of date, that embedding depends only on earlier events. Embeddings from different checkpoints are not comparable, so each version of the backbone needs its own feature set, and a head is valid only on the version it was fitted on.

\paragraph{Delta scoring} The same property lets a daily job rescore only the clients whose input changed and reproduce a full pass exactly (Appendix~\ref{app:delta}).

\section{Tokenization}\label{app:tokens}

\paragraph{Profile event} The four fields \tok{src\_type21}, \tok{src\_type22}, \tok{src\_type31} and \tok{src\_type32} never change within a client: each takes one value per client in all 98,721 clients with transactions. On fold 1, by contrast, \tok{src\_type11} and \tok{src\_type12} change on 13--15\% of consecutive transactions, and the destination fields and the event type on 55--63\%. The four fields are attributes of the client's account. Repeated on every transaction they took 4 of 13 tokens, so we state them once, in a profile event at the time of the client's first event. Transactions shrink to 9 tokens, and the pretraining stream from 454 million to 332 million tokens.

\paragraph{Geolocation} Raw pings outnumber transactions 1.7 to 1 and would crowd them out of a window. We keep one geolocation event per client per Moscow day. It carries the day's most frequent \tok{geohash\_5} cell and the \tok{geohash\_4} cell that contains it, and it is stamped at the day's last ping, so it never runs ahead of the pings it summarizes. This leaves 14.7 million events.

\paragraph{Vocabulary} Values seen in the pretraining folds become tokens. Values of the two counterparty-like fields seen fewer than 1,000 times, and geohash cells seen fewer than 100 (\tok{g4}) or 300 (\tok{g5}) times, fall into one \tok{other} token per key. The vocabulary has 12,828 tokens, and no token of the held-out fold is unknown.

\begin{table}[htbp]
\centering
\caption{Engineered features fitted on $n$ labeled client-months drawn at random, against EventGPT zero-shot (mean of the two seeds of Table~\ref{tab:zero-shot}, no labels). ROC-AUC on the test fold, mean over draws (five up to 30,000 rows, three at 100,000, two at 300,000), with the rule for earlier occurrences applied to both; in brackets the mean number of positives in a draw. The last column is the engineered-feature baseline of Table~\ref{tab:zero-shot}, fitted on every positive of the pretraining folds and 60,000 negatives, under the same rule.}
\label{tab:labels}
\footnotesize
\setlength{\tabcolsep}{3pt}
\begin{tabular}{lcccccccc}
\toprule
 & zero-shot & \multicolumn{6}{c}{engineered features, labeled rows $n$} & all \\
\cmidrule(lr){3-8}
target & (no labels) & 1,000 & 3,000 & 10,000 & 30,000 & 100,000 & 300,000 & labels \\
\midrule
first 48/31 & 0.909 & 0.573 (0) & 0.650 (0) & 0.718 (2) & 0.871 (6) & 0.894 (19) & 0.912 (66) & 0.921 \\
first 23/49 & 0.918 & 0.784 (2) & 0.809 (2) & 0.866 (7) & 0.856 (25) & 0.896 (81) & 0.902 (263) & 0.930 \\
first 26/28 & 0.828 & 0.735 (2) & 0.739 (4) & 0.776 (11) & 0.817 (34) & 0.858 (128) & 0.849 (355) & 0.880 \\
first 48/10 & 0.787 & 0.739 (2) & 0.668 (5) & 0.695 (11) & 0.763 (34) & 0.771 (134) & 0.826 (394) & 0.841 \\
first 50/19 & 0.905 & 0.746 (2) & 0.821 (8) & 0.854 (34) & 0.921 (95) & 0.929 (342) & 0.932 (992) & 0.937 \\
\bottomrule
\end{tabular}
\end{table}

\section{Top of the ranking}\label{app:top}

On PR-AUC the hidden state scores below the engineered features on every official target, by more than its ROC-AUC suggests (Table~\ref{tab:main-pr}). With the profile fields and token counts EventGPT scores above the engineered features on \tok{target\_2} and \tok{target\_4}. On fold 0 the ensemble has the highest PR-AUC and top-1\% recall. Over five folds it ties with the engineered features on top-1\% recall. We also measure recall in the top 0.5\%, 1\% and 5\% of test rows over five folds and two seeds, averaged over the four product targets. The ensemble recalls 0.103, 0.150 and 0.362 against 0.100, 0.150 and 0.358 for the engineered features. It recalls more on \tok{target\_4} in 28 of the 30 fold, seed and cut combinations and less on \tok{target\_1} in the top 0.5\% in all ten. Precision in the same cuts is 0.040, 0.032 and 0.018 for both.

\begin{table}[htbp]
\centering
\small
\setlength{\tabcolsep}{4pt}
\caption{PR-AUC on the official targets, mean of two seeds, and the share of positives in the top 1\% of test rows. A random ranking scores the prevalence and 0.010. Late fusion and the ensemble are those of Table~\ref{tab:gap}.}
\label{tab:main-pr}
\begin{tabular}{lccccc}
\toprule
 & & engineered & \multicolumn{2}{c}{EventGPT} & \\
\cmidrule(lr){4-5}
target & prevalence & features & hidden state & + profile fields, token counts & ensemble \\
\midrule
\tok{target\_1} & 0.0044 & 0.023 & 0.014 & 0.016 & 0.023 \\
\tok{target\_2} & 0.0005 & 0.014 & 0.008 & 0.017 & 0.021 \\
\tok{target\_3} & 0.0038 & 0.021 & 0.014 & 0.019 & 0.022 \\
\tok{target\_4} & 0.0026 & 0.017 & 0.011 & 0.018 & 0.019 \\
mean & 0.0028 & 0.019 & 0.012 & 0.017 & 0.021 \\
\midrule
recall, top 1\% & 0.010 & 0.157 & 0.113 & 0.154 & 0.166 \\
\bottomrule
\end{tabular}
\end{table}

\begin{table}[htbp]
\centering
\caption{EventGPT and the flat decoder, both on windows at random offsets, per pretraining seed. Throughput on one NVIDIA L4; validation loss and top-1 accuracy of the next token at the end of pretraining; the official mean uses the last hidden state for EventGPT and the recency-weighted embedding for the flat decoder. The flat decoder attends to every earlier token, so it can copy fields of recent events, and most of its lower loss is on the geolocation cell; on the event name the two are level. Its official mean is 0.003 above EventGPT's, less than the 0.018 between its own two seeds.}
\label{tab:flat}
\begin{tabular}{lcccc}
\toprule
 & \multicolumn{2}{c}{EventGPT} & \multicolumn{2}{c}{flat decoder} \\
\cmidrule(lr){2-3}\cmidrule(lr){4-5}
 & seed 42 & seed 43 & seed 42 & seed 43 \\
\midrule
throughput (tokens per second) & 117,300 & 117,600 & 47,100 & 55,600 \\
validation loss & 1.023 & 1.020 & 0.944 & 0.947 \\
top-1 accuracy, \tok{geohash\_5} & 0.582 & 0.567 & 0.686 & 0.680 \\
top-1 accuracy, event name & 0.573 & 0.574 & 0.579 & 0.580 \\
official mean ROC-AUC & 0.774 & 0.775 & 0.786 & 0.768 \\
\bottomrule
\end{tabular}
\end{table}

\section{Sources of the gap to engineered features}\label{app:gap}

\paragraph{Static attributes} Without the profile fields the engineered features lose most on \tok{target\_4} (0.036). We add the profile fields to the hidden state by late fusion (Section~\ref{sec:static}). \tok{target\_4} rises from 0.754 to 0.821 (seed 42). With early fusion as well it reaches 0.830 and 0.822 against 0.818 for the engineered features ($p = 0.11$ and 0.61). We also pretrain with early fusion. The hidden state then scores 0.789 against 0.787 without it.

\paragraph{Timing} We add five timing features to the hidden state: the days since the first and the last event and the event counts over 30 days, 90 days and all history. They raise the official mean by 0.001 and 0.003 and leave churn unchanged. The engineered features lose 0.009 without their own timing features, so the hidden state already carries this information.

\paragraph{History before the window} We add the full-history token counts to the hidden state and the profile fields. They raise churn, first 50/19 (from about 0.84 to 0.92) and first 48/10 (from about 0.80 to 0.83) on both seeds. Token counts over EventGPT's window score 0.026 below token counts over the full history. A longer window did not recover this history for EventGPT (Section~\ref{sec:context}).

\paragraph{Schema knowledge} The engineered features score 0.816 and full-history token counts 0.803, so knowledge of the schema adds 0.013.

\paragraph{Amount resolution} EventGPT sees each amount as one of 21 quantile bins. We replace each amount in the engineered features by the mean of its bin. The official mean falls by 0.005 with 21 bins and does not change with 256 bins.

\section{Ablations}\label{sec:ablations}

We change one element of the model or its data at a time (Table~\ref{tab:ablations}). We run two seeds except where marked.

\begin{table}[htbp]
\centering
\caption{Mean ROC-AUC over the official targets, logistic head. Rows with two seeds give the mean and, in brackets, the two seeds' scores; the row marked $^\ddagger$ uses transactions only; the row marked $^\S$ trains at learning rate $3\times10^{-4}$. Every row but the windows and width rows uses windows at random offsets; the width rows add QK normalization. The pretraining-period rows compare with transactions only over the full period (0.793), and every other row with the first.}
\label{tab:ablations}
\footnotesize
\begin{tabular}{llcl}
\toprule
element & variant & seeds & mean ROC-AUC \\
\midrule
architecture & hierarchy (EventGPT), last state & 2 & 0.774 (0.774, 0.775) \\
 & flat decoder, last token & 2 & 0.714 (0.713, 0.714) \\
 & flat decoder, recency over tokens & 2 & 0.777 (0.768, 0.786) \\
positions$^\S$ & temporal RoPE & 2 & 0.781 (0.777, 0.786) \\
 & ordinal & 2 & 0.778 (0.775, 0.780) \\
modalities & transactions + geolocation & 2 & 0.774 (0.774, 0.775) \\
 & transactions only & 2 & 0.793 (0.791, 0.795) \\
pretraining period$^\ddagger$ & before Feb.~2022, same steps & 2 & 0.789 (0.788, 0.790) \\
 & before Feb.~2022, same passes per token & 2 & 0.787 (0.784, 0.791) \\
training loss & without \tok{dt}, \tok{amt\_bin}, \tok{g4}, \tok{g5} & 1 & 0.775 \\
embedding & mean & 2 & 0.755 (0.752, 0.758) \\
 & recency & 2 & 0.765 (0.761, 0.769) \\
training windows & aligned, with QK normalization & 2 & 0.787 (0.786, 0.788) \\
attention & QK normalization & 1 & 0.779 \\
width & 384 / 512 / 1024, aligned & 1 & 0.786 / 0.791 / 0.791 \\
event encoder & one feed-forward per event kind & 1 & 0.779 \\
\bottomrule
\end{tabular}
\end{table}

We pretrain and evaluate on transactions alone. The official mean rises from 0.774 to 0.793 and churn from 0.934 to 0.950 on both seeds, and the classification head rises on all five event targets. Without geolocation the engineered features lose 0.001 while EventGPT gains 0.019. Without daily geolocation events the 2,048-token window holds more transactions. Token counts show the same effect: their loss to the window shrinks from 0.026 to 0.008. Within the transactions-only window EventGPT scores above token counts and level with token counts over the full transaction history. Every other result here uses geolocation, as MBD does.

Without geolocation the zero-shot scores fall to 0.72 to 0.89, from 0.79 to 0.92. The label asks whether the kind occurs within the month, which also depends on how many events the month holds. With a geolocation event on most days, a client who transacts rarely has a lower score for every transaction kind. The score then carries the client's activity, which the monthly label rewards. We correlate the log score with the log of the client's recent transactions: 0.39 to 0.49 with geolocation and $-0.15$ to 0.20 without (seed 42, as in the next paragraph).

We multiply the score by one plus the client's transactions in the 30 days before the as-of date. This uses no labels. The scaled score reaches 0.79 to 0.93 on transactions alone and 0.82 to 0.93 with geolocation, above the count alone. We first tried this scaling on the test fold, after the scores of Table~\ref{tab:zero-shot} were known, beside a second form based on transactions per day over the whole history. We then compared the two on training fold 1. The 30-day count scores above the daily rate for every model and target and raises every unscaled score. Table~\ref{tab:zero-shot} keeps the unscaled scores, which use only the model.

We remove the loss on gaps, amount bins and geohash cells, the tokens a product target is least likely to need. We keep these tokens in the input and drop only their prediction. The official mean does not change (0.775 against 0.774). Churn rises slightly (seed 42), and the zero-shot scores change by at most 0.023.

We replace temporal RoPE with learned ordinal positions in the history decoder. At the learning rate of the other rows, $6\times10^{-4}$, the ordinal model diverges after 3,000 to 5,000 steps in both seeds. At $3\times10^{-4}$ both train stably. Temporal RoPE scores 0.781 and ordinal positions 0.778. Temporal RoPE scores higher on both seeds, but the mean difference, 0.003, is smaller than the difference between two seeds of either model (0.009 and 0.005). The ordinal model also adds learned positions within each event, which the RoPE model leaves out, so the comparison does not isolate the position scheme.

We also read the history decoder's attention as weights over a client's events. We average the heads and roll the attention out through the six layers \cite{rollout}, with the residual path weighted one half. We pool the hidden states with the last event's rolled-out weights. On 2,000 test histories the weights spread over a median of 102 of 219 events (the exponential of their entropy) and put 2.4\% on the last event, so the pooled vector is close to the mean. Averaged over two seeds it scores 0.760 (0.757 and 0.763), between mean pooling (0.757) and recency pooling (0.765) and 0.015 below the last hidden state (0.774). We also learn the pooling: an attention head over the frozen hidden states of the last 128 events, with one learned query per label, fitted on 60,000 training rows. On aligned windows it scores 0.718 and 0.727 on the two seeds, against 0.784 and 0.790 for the last hidden state on the same rows.

We test query-key normalization and one feed-forward network per event kind in the event encoder (transactions, geolocation and the rest, routed by the event name). Each adds 0.005 on seed 42. Two seeds of the same configuration differ by up to 0.018 for the flat decoder and up to 0.022 for EventGPT across the five test folds, so a difference of 0.005 on one seed is not established.

We test width 512 (16 heads of width 32, 38.4 million parameters) and width 1024 (16 heads of width 64, 139.7 million) on aligned windows with query-key normalization (seed 42). Both lower the validation loss, to 0.997 at width 1024 from 1.020, and raise the official mean from 0.786 to 0.791. With late fusion of the profile fields the gain disappears. At width 1024 late fusion lowers the official mean, while it raises it at the smaller widths. The four field embeddings add 4,096 columns to a head regularized as at width 384, which may explain this.

\section{Length of history}\label{sec:context}

We train EventGPT and the flat decoder on windows of 4,096 and 8,192 tokens (Table~\ref{tab:window}). MBD histories are long: the 90th percentile client has about 1,000 transactions and 300 days of geolocation, while a 2,048-token window holds about 250 events. Token counts gain 0.024 from 2,048 to 8,192 tokens. EventGPT stays level on random windows (0.774 to 0.772) and loses 0.008 on aligned windows (0.788 to 0.780). The flat decoder stays level at 4,096 tokens. Churn does not change with the window for any model. The extra history carries information for the product targets, and EventGPT's embedding does not take it up at this size. Pooled readouts do not take it up either: at 8,192 tokens mean and recency pooling score 0.748 and 0.762 on random windows and 0.743 and 0.765 on aligned windows, below the last hidden state.

EventGPT trains at about 126,000 tokens per second at 8,192 tokens against 118,000 at 2,048. Its history decoder attends over events, and with block attention its token stages grow linearly with the window. The flat decoder falls from 47,000 tokens per second at 2,048 tokens to about 30,000 at 4,096. We did not train it at 8,192 tokens: at 4,096 a run already takes about six hours on the L4, and its attention cost grows with the square of the window.

\paragraph{Block attention within events}\label{sec:packed}

Neither the event encoder nor the token decoder attends across an event boundary (Figure~\ref{fig:masks}), but a dense attention mask over a window of $t$ tokens still costs memory in proportion to $t^2$. We lay each event out instead as one row of at most $L$ slots, with $L$ the longest event in the window (ten tokens on MBD), and attend within rows. Both stages then cost in proportion to the number of events times $L^2$. On real windows the block and the dense computations agree to within $1.2 \times 10^{-6}$ on every event. Block attention makes windows of 8,192 tokens affordable.

\begin{table}[htbp]
\centering
\caption{Training time on one NVIDIA L4 GPU for 20,000 steps of 32,768 tokens (655 million tokens), windows at random offsets, one run per cell, without query-key normalization. Throughput is in thousands of tokens per second, from the training log; for runs resumed after a preemption it is measured within each uninterrupted segment. Query-key normalization slows EventGPT to about 88,000 tokens per second (2.1 hours at 2,048 tokens, aligned windows).}
\label{tab:gpu}
\begin{tabular}{lcccc}
\toprule
 & \multicolumn{2}{c}{EventGPT} & \multicolumn{2}{c}{flat decoder} \\
window & hours & tokens/s & hours & tokens/s \\
\midrule
2,048 & 1.5 & 118k & 3.9 & 47k \\
4,096 & 1.5 & 125k & 6.2 & 29k \\
8,192 & 1.4 & 126k & & \\
\bottomrule
\end{tabular}
\end{table}

\begin{table}[htbp]
\centering
\caption{Mean ROC-AUC over the official targets (churn without January 2023 in parentheses) against the window, seed 42 for the models. The aligned column is trained without query-key normalization (0.786 with it at 2,048 tokens). Token counts at full history reach 0.803.}
\label{tab:window}
\begin{tabular}{lcccc}
\toprule
 & token counts & \multicolumn{2}{c}{EventGPT} & flat decoder \\
window & (CatBoost) & random & aligned & recency \\
\midrule
2,048 & 0.777 & 0.774 (0.934) & 0.788 (0.933) & 0.786 (0.900) \\
4,096 & 0.782 & 0.772 (0.936) & & 0.785 (0.896) \\
8,192 & 0.801 & 0.772 (0.935) & 0.780 (0.935) & \\
\bottomrule
\end{tabular}
\end{table}

\begin{figure}[htbp]
\centering
\includegraphics[width=\textwidth]{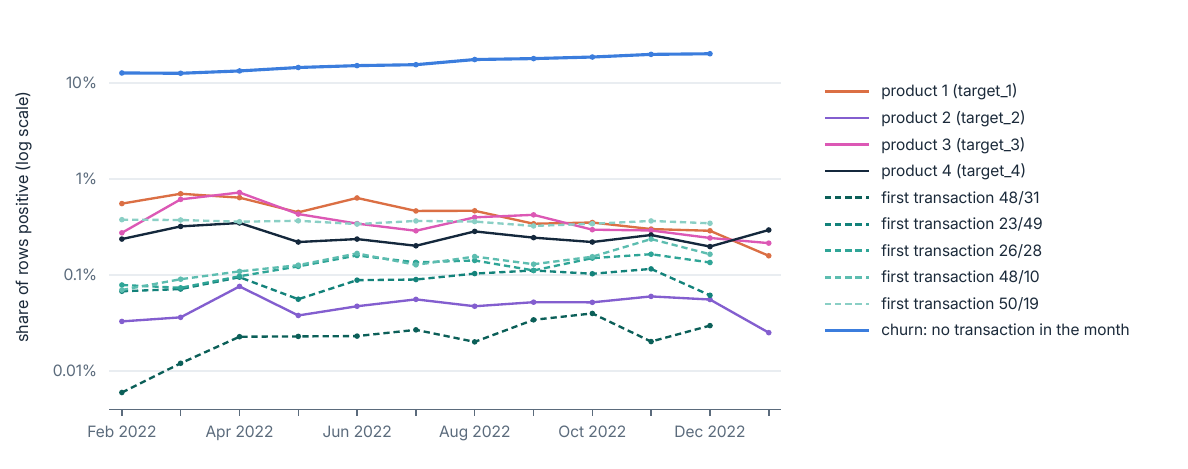}
\caption{Share of rows positive for each label by month, on a log scale, over all five folds (the test fold in full, the training rows weighted back to their share). Product targets are solid, event targets dashed and churn blue. January 2023 follows the last transaction, so the event targets and churn end in December.}
\label{fig:labels}
\end{figure}


\begin{thebibliography}{99}
\raggedright

\bibitem{rollout} Abnar, S., and Zuidema, W. (2020). Quantifying attention flow in transformers. In \emph{Proceedings of the 58th Annual Meeting of the Association for Computational Linguistics (ACL)}, 4190--4197.

\bibitem{coles} Babaev, D., Ovsov, N., Kireev, I., Ivanova, M., Gusev, G., Nazarov, I., and Tuzhilin, A. (2022). CoLES: Contrastive learning for event sequences with self-supervision. In \emph{Proceedings of the 2022 International Conference on Management of Data (SIGMOD)}. doi:10.1145/3514221.3526129.

\bibitem{timesformer} Bertasius, G., Wang, H., and Torresani, L. (2021). Is space-time attention all you need for video understanding? In \emph{Proceedings of the 38th International Conference on Machine Learning (ICML)}. arXiv:2102.05095.

\bibitem{nuformer} Braithwaite, D.~T., Cavalcanti, M., McEver, R.~A., Udagawa, H., Silva, D., Ramanath, R., Meneses, F., Yoshida, A., Wingert, E., Ramos, M., Zanfelice, B., and Gupta, A. (2025). Your spending needs attention: Modeling financial habits with transformers. arXiv:2507.23267.

\bibitem{delong} DeLong, E.~R., DeLong, D.~M., and Clarke-Pearson, D.~L. (1988). Comparing the areas under two or more correlated receiver operating characteristic curves: a nonparametric approach. \emph{Biometrics} 44(3), 837--845.

\bibitem{transactiongpt} Dou, Y., Jiang, Z., Zhang, T., Hu, M., Xu, Z., Jain, S., Saini, U.~S., et al. (2025). TransactionGPT. arXiv:2511.08939.

\bibitem{latte} Fadeev, E., Mollaev, D., Shestov, A., Zoloev, O., Sakhno, A., Korolev, D., Kireev, I., Savchenko, A., and Makarenko, M. (2025). LATTE: Learning aligned transactions and textual embeddings for bank clients. arXiv:2508.10021.

\bibitem{qknorm} Henry, A., Dachapally, P.~R., Pawar, S., and Chen, Y. (2020). Query-key normalization for transformers. In \emph{Findings of the Association for Computational Linguistics: EMNLP 2020}, 4246--4253.

\bibitem{eventgptcam} Liu, S., Li, J., Zhao, G., Zhang, Y., Meng, X., Yu, F.~R., Ji, X., and Li, M. (2025). EventGPT: Event stream understanding with multimodal large language models. In \emph{Proceedings of the IEEE/CVF Conference on Computer Vision and Pattern Recognition (CVPR)}. doi:10.1109/CVPR52734.2025.02713.

\bibitem{mbdmini} MBD-mini. Hugging Face dataset \tok{ai-lab/MBD-mini}, CC BY 4.0.
\url{https://huggingface.co/datasets/ai-lab/MBD-mini}.

\bibitem{esgpt} McDermott, M., Nestor, B., Argaw, P., and Kohane, I. (2023). Event Stream GPT: A data pre-processing and modeling library for generative, pre-trained transformers over continuous-time sequences of complex events. In \emph{Advances in Neural Information Processing Systems 36, Datasets and Benchmarks Track}.

\bibitem{mbd} Mollaev, D., Kostin, A., Postnova, M., Karpukhin, I., Kireev, I., Gusev, G., and Savchenko, A. (2024). Multimodal Banking Dataset: Understanding client needs through event sequences. arXiv:2409.17587.

\bibitem{nvidia} NVIDIA. Build your own transaction foundation model for financial intelligence.
\url{https://developer.nvidia.com/blog/build-your-own-transaction-foundation-model-for-financial-intelligence/}.

\bibitem{obuchowski} Obuchowski, N.~A. (1997). Nonparametric analysis of clustered ROC curve data.
\emph{Biometrics} 53(2), 567--578.

\bibitem{pragma} Ostroukhov, M., Mikhailov, R., Iashin, V., Sokolov, A., Akshonov, A., Protasov, V.,
Goncharov, A., Beloborodov, D., Mullin, V., Enzmann, R.~Y., Kolovos, G., Renders, J., Nesterov, P., and Repushko, A. (2026). PRAGMA: Revolut foundation model. arXiv:2604.08649.

\bibitem{tabbert} Padhi, I., Schiff, Y., Melnyk, I., Rigotti, M., Mroueh, Y., Dognin, P., Ross, J., Nair, R., and Altman, E. (2021). Tabular transformers for modeling multivariate time series. In \emph{ICASSP 2021}. arXiv:2011.01843.

\bibitem{openbanking} Polleti, G., Santana, M., and Fontes, E. (2025). Open Banking foundational model: Learning language representations from few financial transactions. arXiv:2511.12154.

\bibitem{catboost} Prokhorenkova, L., Gusev, G., Vorobev, A., Dorogush, A.~V., and Gulin, A. (2018).
CatBoost: unbiased boosting with categorical features. In \emph{Advances in Neural Information
Processing Systems 31}.

\bibitem{lifestream} Sakhno, A., et al. (2025). PyTorch-Lifestream: Learning embeddings on discrete event sequences. In \emph{Proceedings of the 34th International Joint Conference on Artificial Intelligence (IJCAI), Demo Track}, 11104--11108. doi:10.24963/ijcai.2025/1272.
\url{https://github.com/dllllb/pytorch-lifestream}.

\bibitem{starling} Starling Bank (2026). Starling backs UK sovereign cloud. Press release, 14 October 2026. \url{https://www.starlingbank.com/news/starling-backs-uk-sovereign-cloud/}.

\bibitem{rope} Su, J., Lu, Y., Pan, S., Murtadha, A., Wen, B., and Liu, Y. (2021). RoFormer: Enhanced transformer with rotary position embedding. arXiv:2104.09864.

\bibitem{qwenomni} Xu, J., Guo, Z., He, J., Hu, H., et al. (2025). Qwen2.5-Omni technical report. arXiv:2503.20215.

\bibitem{thp} Zuo, S., Jiang, H., Li, Z., Zhao, T., and Zha, H. (2020). Transformer Hawkes process. In \emph{Proceedings of the 37th International Conference on Machine Learning (ICML)}, PMLR 119, 11692--11702.

\end{thebibliography}
\end{document}